\documentclass[10pt,twocolumn,letterpaper]{article}

\usepackage[pagenumbers]{cvpr} % To force page numbers, e.g. for an arXiv version

\usepackage{amsmath}
\usepackage{amssymb}
\usepackage[table]{xcolor}

\newcommand{\early}{\textsc{GMC-H2}}
\newcommand{\faithful}{\textsc{GMC-L16}}
\newcommand{\R}{\mathbb{R}}
\newcommand{\norm}[1]{\left\lVert #1\right\rVert}
\newcommand{\yes}{\checkmark}
\newcommand{\no}{\ensuremath{\times}}

\definecolor{cvprblue}{rgb}{0.21,0.49,0.74}
\usepackage[pagebackref,breaklinks,colorlinks,allcolors=cvprblue]{hyperref}

\title{Messages, Not Tokens: Grounded Coresets for Faithful VLM Compression}

\author{
Long Qian$^{1,2\dag}$ \quad Jiaqi Wei$^{4\dag}$ \quad Bingke Zhu$^{1,2*}$ \quad Yingying Chen$^{1,2}$ \quad Jinqiao Wang$^{1,2,3}$\\
$^1$Foundation Model Research Center, Institute of Automation, Chinese Academy of Sciences, Beijing, China\\
$^2$School of Future Technology, University of Chinese Academy of Sciences, Beijing, China\\
$^3$Wuhan AI Research, Wuhan, China\\
$^4$School of Engineering, Cardiff University, Cardiff, United Kingdom\\
{\tt\small qianlong2024@ia.ac.cn, WeiJ16@cardiff.ac.uk, \{bingke.zhu,yingying.chen,jqwang\}@nlpr.ia.ac.cn}
}

\begin{document}
\maketitle
{\let\thefootnote\relax
\footnotetext{$\dag$~Equal contribution.}
\footnotetext{*~Corresponding author.}}
\begin{abstract}
Modern vision language models (VLMs) turn high-resolution images into long sequences of visual tokens. Every token traverses the language decoder and
persists in its prompt KV cache, inflating inference cost and motivating aggressive
visual compression. Existing score-based methods assign each token an independent
importance score and retain the Top-$K$. However, text queries consume collective, signed attention
messages from the visual population, not isolated patches. Consequently, equally
sized Top-$K$ sets can repeatedly cover one salient region, omit sparse but
complementary evidence and discard
information carried by the removed population. We therefore formulate faithful visual compression as constructing a compact coreset for decoder messages, and introduce our training-free \emph{Grounded Message Coreset
Pruning} (GMC) which jointly allocates support across query-grounded, appearance, and
coordinate-aware evidence, then transports discarded states into selected
representatives at their original multimodal positions before physical compaction
and native attention resume. This decomposes faithful compression into two coupled components, including selecting
carriers that cover the required message modes and realizing the signed population
message on those carriers. We further derive bounds connecting their errors to
signed-message distortion, visual innovation, and candidate-margin stability.
Experiments across multiple VLM families and diverse benchmarks demonstrate
strong performance, with GMC-H2 retaining 97.78\% Full-relative mean capability
on Qwen2.5-VL-7B using 80.2\% fewer visual tokens, while GMC-L16 reaches
100.36\%. Controlled interventions verify that collective support and
population realization jointly drive these gains.
\end{abstract}
\section{Introduction}

Modern VLMs encode high-resolution documents, charts, and natural scenes into long visual prefixes~\citep{liu2023llava,bai2025qwen25vl}. Every patch is
projected through the language decoder, enlarges prefill attention, and occupies
prompt KV cache throughout generation. Compression is therefore not merely an
input preprocessing choice, it determines how much visual evidence remains
available to every subsequent reasoning step~\citep{chen2024fastv,yang2025visionzip,
zhang2025vispruner,zhu2026hawk}.

\begin{figure}[!t]
  \centering
  \includegraphics[width=\columnwidth]{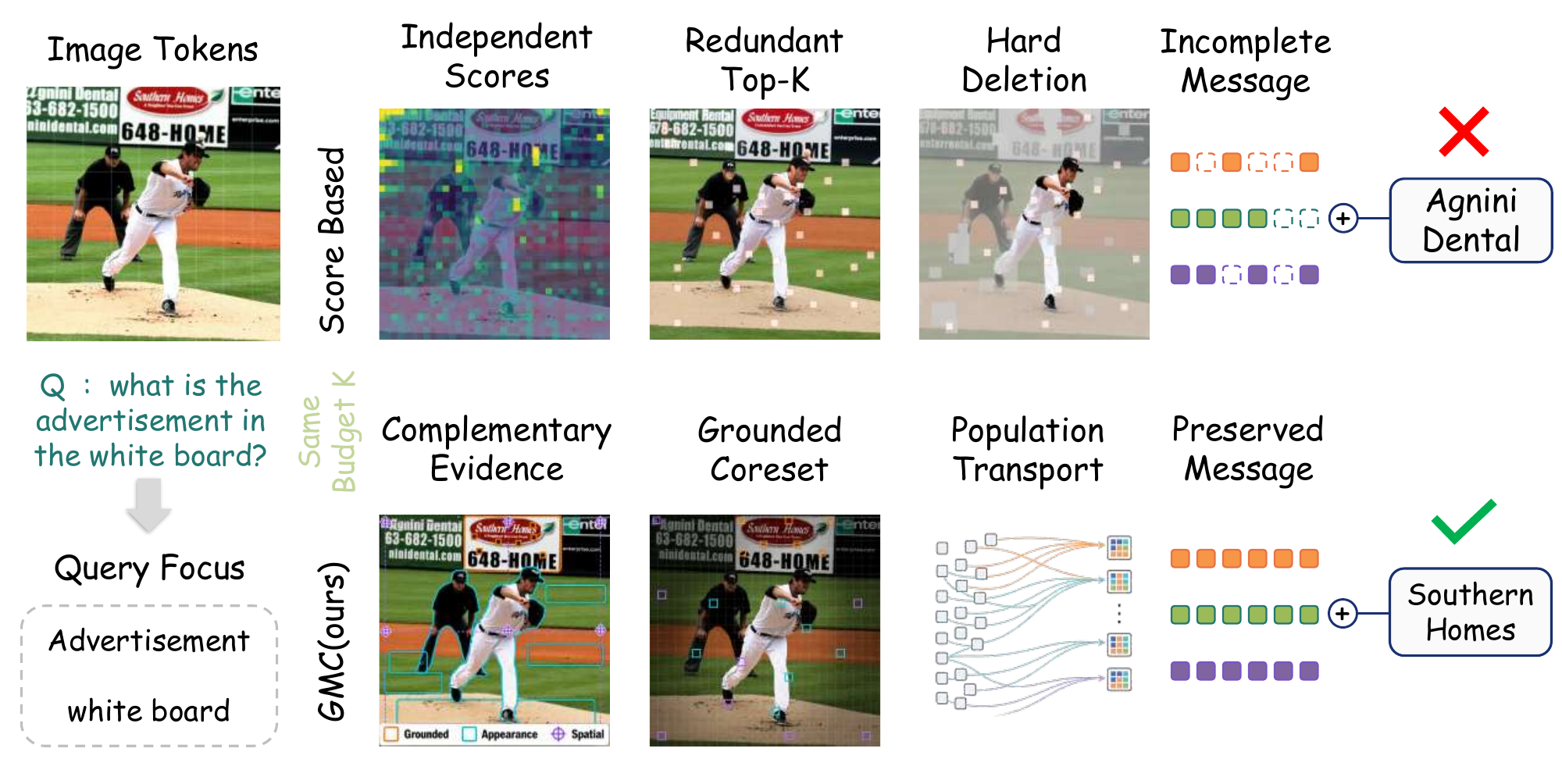}
  \caption{\textbf{Why equal-$K$ token sets carry different evidence}. Independent
  Top-$K$ concentrates on individually salient patches, leaving complementary
  message carriers uncovered and breaking the aggregate consumed downstream.
  Our method GMC jointly covers query-grounded, appearance, and spatial evidence, then
  transports the discarded population into the selected representatives.}
  \label{fig:motivation}
\end{figure}

Most methods reduce compression to a scalar question: \emph{how important is
token $i$?} Attention, norm, or sensitivity scores retain the largest values,
yet top-ranked patches may redundantly describe one object while losing a
numeral, axis tick, or relation endpoint. Diversity objectives reduce
duplication, but raw-feature coverage does not guarantee that the selected set
spans the evidence required by the decoder. This limitation is particularly
consequential because text queries do not consume visual tokens independently;
instead, they aggregate contributions from the visual population into a
collective, query-conditioned message. The evidence needed for an answer may
therefore be distributed across multiple complementary patches: a word may span
several regions, a chart answer may require a bar, an axis, and a tick, and a
spatial answer may require two objects together with their coordinate relation.
Hard deletion removes the information carried by the residual population, while
merging after an independently chosen support decouples \emph{where} evidence is
retained from \emph{what} each representative should carry. Faithful compression
must instead preserve complementary message carriers together with the
information contributed by discarded visual tokens.

We formulate these requirements as a \emph{grounded message coreset}. Its central
claim is beyond a new token score: support allocation and message realization
form two coupled components. At a boundary, our method uses native head query pairs to induce
distributions over message-bearing carriers, appearance and spatial clients protect
future content and geometry not yet activated by the prompt. Batched responsibility
allocates complementary support. Population transport then moves discarded states
to that support, preserves original multimodal coordinates, and lets native compact
attention realize the signed message consumed downstream.

Figure~\ref{fig:motivation} contrasts these outcomes under the same budget. The
independent branch concentrates support around one high-score region and leaves an
incomplete aggregate; the coreset branch distributes carriers across three
complementary evidence families and realizes discarded states.
Faithful compression therefore requires a jointly chosen support, a population
realization map, and the original coordinate system on which compact attention is
recomputed.

This view makes Full-relative fidelity auditable: coverage bounds uncovered client
modes, transport maps their deficit to boundary-state error, a signed-message
identity separates state error from compact-attention mass mismatch, and centered
logit distortion controls represented candidate ordering. Experiments intervene on
each component, trace its error to answer flips, test factual grounding under both
discriminative and generative protocols, and verify physical compaction across two
decoder families.

Our contributions are:
\begin{itemize}
  \item We revisit visual token compression and formulate it as a
  \emph{grounded message coreset} problem that jointly determines where
  complementary evidence is retained and what each representative carries.

  \item We propose \emph{Grounded Message Coreset Pruning} (GMC), a
  training-free framework that allocates support across query-grounded,
  appearance, and spatial evidence, and transports discarded states into
  selected representatives while preserving their original multimodal
  positions.

  \item Extensive experiments across multiple VLM families and diverse
  benchmarks show that GMC achieves state-of-the-art performance under severe
  token budgets. Controlled studies further validate both complementary support
  allocation and population realization.
\end{itemize}

\section{Related Work}

\paragraph{Token importance and structure.}
Training-free methods rank visual tokens by text attention, feature norm, or
redundancy~\citep{chen2024fastv,yang2025visionzip,ye2025fitprune,
zhang2024sparsevlm}. Recent work calibrates head importance, corrects positional
bias, estimates perturbation sensitivity, follows information flow, or balances
relevance with diversity~\citep{zhu2026hawk,zhang2026d2pruner,kim2026zooprune,
sun2026ifprune,wang2026posprune}. These advances produce stronger scalar or
pairwise criteria, yet the retained set is still commonly determined before the
downstream representation of deleted tokens is specified.

\paragraph{Coverage and recovery.}
MMTok and CoIn move from independent ranking toward multimodal coverage and
informative set selection~\citep{dong2026mmtok,du2026coin}; DART, DivPrune, VisPruner, and
VLM-Pruner reduce duplication while preserving visual or spatial structure
~\citep{wen2025dart,alvar2025divprune,zhang2025vispruner,wu2026vlmpruner}. Hi-Lo Prune, ApET,
VFlowOpt, and related compression methods recover, recycle, or approximate the
information that hard deletion would remove~\citep{sun2026hilo,ma2026apet,
yang2025vflowopt}.

\paragraph{Collective coresets.}
Location supplies diminishing-return coverage; weighted
coresets and token merging instead represent a population by compact carriers
~\citep{nemhauser1978analysis,bachem2018scalable,bolya2023tome}. At a VLM boundary, coverage allocates carriers by nonnegative responsibility,
while transport and native compact softmax realize the signed decoder message.
This distinction matters because equal-norm contributions may cancel when
oppositely directed.

\begin{figure*}[!t]
  \centering
  \includegraphics[width=\textwidth]{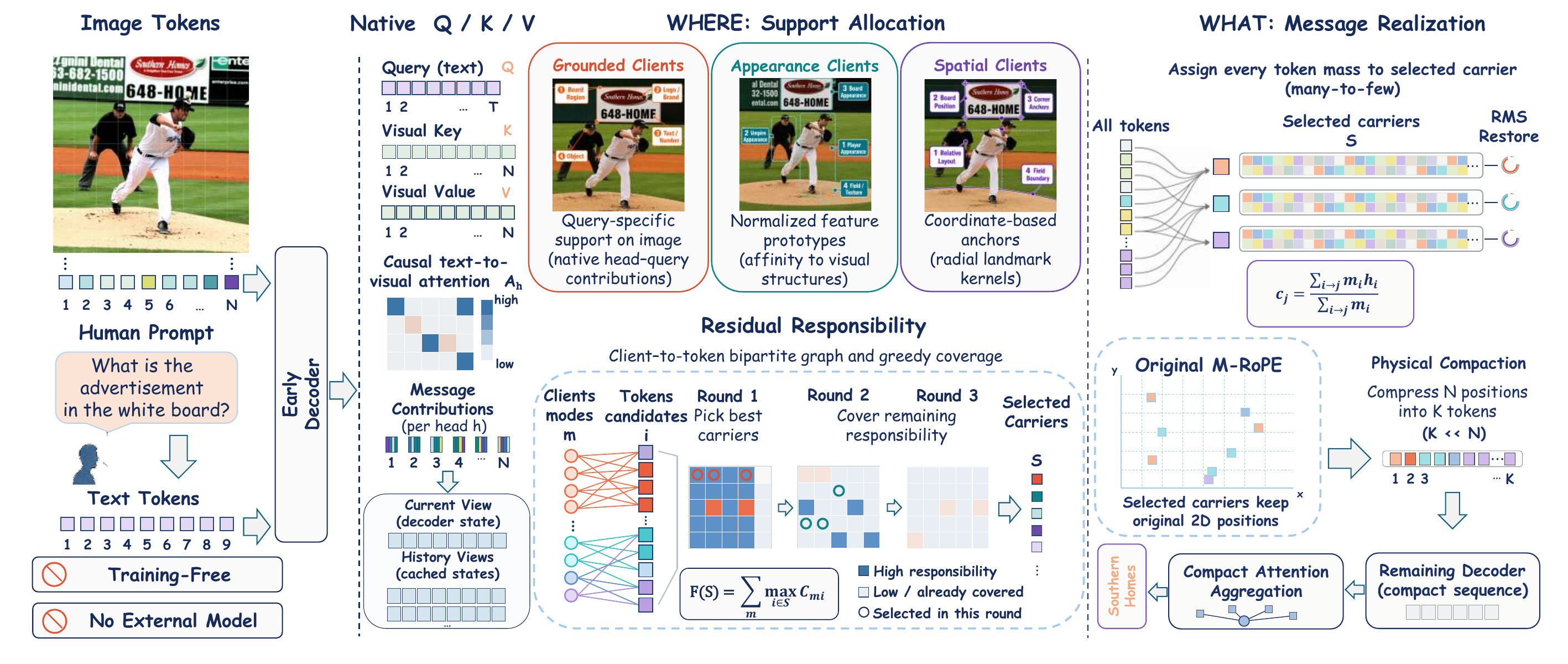}
  \caption{\textbf{Overview of GMC}. \emph{Support allocation} forms
  grounded-message clients from native head query contributions and augments
  them with appearance and spatial clients; coverage-driven selection assigns
  the finite budget to complementary carriers rather than isolated high scores.
  \emph{Message realization} transports every discarded state into those
  representatives, retains their original multimodal coordinates, and resumes
  native compact attention on the resulting aggregate decoder message.}
  \label{fig:method}
\end{figure*}

\section{Methodology}
\label{sec:method}

\subsection{Problem Formulation}

Figure~\ref{fig:method} demonstrates our method. The three banks determine \emph{where} a finite
support should be allocated; population transport determines \emph{what} each
selected carries when native compact attention recomputes the
signed message.

Let $H_V=(h_1,\ldots,h_N)$ denote visual states at boundary $p$, $q$ the
prompt, and $I$ the number of images. A requested budget $K_r$ becomes
$K=\min\{N,\max\{K_r,I\}\}$, guaranteeing one carrier per image and identity
when no compression is requested. GMC jointly returns $S\subseteq[N]$,
$|S|=K$, and representatives $\widetilde H_S$ to minimize the downstream
distortion $D(f_{p:L}(q,H_V),f_{p:L}(q,\widetilde H_S))$ under the compute cost
$C(p,K)$. Decoding every candidate support is combinatorial; GMC instead
optimizes a tractable carrier-allocation interface, reconstructs the discarded
population, and audits the signed message after compact attention is recomputed.

\subsection{Boundary Message Geometry}

Before decoder layer $p$, native RMS normalization and M-RoPE produce
$Q_h=X_{\mathcal T}W_h^Q$, $K_h^{\rm all}=X_{\le\mathcal T}W_h^K$, and
$V_h=X_{\mathcal V}W_h^V$. We use the visual-column slice $A_h$ of the
\emph{native} causal softmax, so $Y_h=A_hV_h$ includes competition with all
preceding nonvisual keys rather than a visual-only renormalization. Let $O_h$
be head $h$'s output-projection block and $P\in\R^{d\times r}$ the fixed
rank-$r$ orthonormal probe ($r=4$). Writing
$Z_{c,h}=(V_{c,h}O_h)P\in\R^{N\times r}$, we append the headwise homogeneous
coordinate
\begin{equation}
 \begin{aligned}
 \sigma_{c,h}
 &=\sqrt{\lambda_{\rm mass}}\,
 \max\!\left\{\norm{Z_{c,h}}_F/\sqrt{Nr},\varepsilon_{\rm mass}\right\},\\[-2pt]
 \lambda_{\rm mass}&=1,\qquad \varepsilon_{\rm mass}=10^{-8}.
 \end{aligned}
 \label{eq:mass-coordinate}
\end{equation}
It is constant across tokens within component $c$ and head $h$, so its
attention-scaled contribution records visual softmax mass on the same
headwise scale as projected value coordinates.

For component $c$
(the current boundary or a retained historical view), query $t$, and token $i$,
define
\begin{align}
 u^{c,h,t}_i&=A_{c,h}[t,i]
 \left[(V_{c,h}[i]O_h)P;\,\sigma_{c,h}\right],\notag\\
P^G_{c,h,t}(i)&=\frac{\norm{u^{c,h,t}_i}_2}
{\sum_j\norm{u^{c,h,t}_j}_2}\quad\text{for an active row}.
\label{eq:grounded-client}
\end{align}
The norm makes $P^G$ a nonnegative carrier-responsibility distribution for set
allocation; transport and native compact attention separately realize the signed
sum. Every positive-energy head query row is normalized, while a zero-energy
row carries no mass. Cross-query transport uses the signed vectors
$r_i^{c,h,t}=\sqrt{\nu_t}u_i^{c,h,t}$; normalized $\nu_t$ emphasizes the
prompt-terminal decision. Equal-mass farthest-first head clusters prevent broad
head families from suppressing specialized heads. Fixed component weights balance current and historical views;
the appendix tabulates every frozen probe, query, and bank setting.

Prompt-conditioned clients describe evidence already accessed by text queries,
but future answer tokens may require details not yet activated. GMC therefore
adds appearance clients $P_m^V(i)\propto\exp(x_m^\top x_i/\tau_v)$ over normalized
boundary features and per-image spatial clients
$P_m^S(i)\propto\exp(-\norm{p_i-\ell_m}_2^2/\tau_s)$ around original-grid
landmarks. Every token is an appearance client in the fixed protocol; a
deterministic bounded bank is used for documents. For bank $b$, active rows are
preweighted as $\bar C_{mi}=(\beta_b/M_b)a_mP_m(i)$, preventing row-count
domination. Its attainable mass $Z_b=F_b(V)$ induces an effective share
$\omega_b=Z_b/\sum_cZ_c$ and the exact audit
$F(S)/F(V)=\sum_b\omega_b[F_b(S)/Z_b]$. Complete normalization, inactive-row,
and bank-mass conventions are tabulated in the appendix for exact reproduction.

\subsection{Complementary Coverage by Responsibility}

GMC maximizes weighted facility-location coverage:
\begin{equation}
 F(S)=\sum_{m=1}^{M}\max_{i\in S}\bar C_{mi},
 \qquad |S|\le K,\quad \bar C_{mi}\ge0.
 \label{eq:coverage}
\end{equation}
We adopt $F(\varnothing)=0$. Unlike top-$K$ scoring, the marginal
$\Delta(i\mid S)=\sum_m[\bar C_{mi}-\max_{j\in S}\bar C_{mj}]_+$ falls once a
client's evidence is represented, so redundant high-score patches immediately lose
marginal value.

Exact sequential greedy is unnecessary for deployment. Let $b_m(S)=\max_{j\in S}\bar C_{mj}$ and $R_{mi}=[\bar C_{mi}-b_m(S)]_+$. A round of size $b$ computes $u_i=\sum_mR_{mi}$, forms a bounded top-gain pool $\mathcal P$, and assigns residual clients by
\begin{align}
 \rho_{mi}&=\operatorname{softmax}_{i\in\mathcal P}
 \left(\frac{R_{mi}}{\tau_r\max_{j\in\mathcal P}R_{mj}+\epsilon}\right),\notag\\
 s_i&=\sum_m\rho_{mi}R_{mi}.
 \label{eq:responsibility}
\end{align}
The top $b$ pool scores enter $S$ together before $b_m(S)$ is updated. Rounds continue to $|S|=K$, each
recomputing responsibility from client mass unexplained by the current support.
For a multi-image prompt, $S$ starts with the maximum-singleton-gain token from
each image, so every image-compatible support $S_i$ below is nonempty;
single-image runs initialize $S_0=\varnothing$ as the empty anchor set.

\subsection{Population Transport}

Selection identifies representative locations; transport determines what those
locations carry. Normalize $h_i$ to $\hat h_i$, let $p_i$ be its original-grid
coordinate, and restrict candidates $S_i$ to the same image. The gap $\delta_i$
between token $i$'s best and second-best cosine matches defines a sample
separability gate $s_x$ and local ambiguity $g_i$. We assign $i$ to
$\pi(i)=\arg\max_{j\in S_i}\{\hat h_i^\top\hat h_j+\lambda_ps_xg_i
\exp[-\norm{p_i-p_j}^2/\tau_s]\}$. Thus clear semantic matches remain global,
whereas ambiguous local details prefer nearby representatives. Frozen gate forms,
edge-case handling, and temperatures are tabulated in the appendix for exact
reproduction.

Let $\bar e_i$ be cluster-balanced grounded-message mass normalized to mean one;
if its normalizer vanishes, we use the uniform fallback $\bar e_i=1$. The deployed
population weight $m_i=(1-\alpha s_x)+\alpha s_x\bar e_i$ is strictly positive.
With $\mathcal C_j=\{i:\pi(i)=j\}$, the population centroid is
\begin{equation}
 \begin{aligned}
 M_j&=\sum_{i\in\mathcal C_j}m_i,\qquad
 c_j=M_j^{-1}\!\sum_{i\in\mathcal C_j}m_i h_i,\quad M_j>0,\\
 \widetilde h_j&=\mathcal R_j\!\left((1-\tau_j)h_j+\tau_jc_j\right),
 \end{aligned}
 \label{eq:transport}
\end{equation}
where $0\le\tau_j\le1$ and $\mathcal R_j(u)=u\,\operatorname{rms}(h_j)/(\operatorname{rms}(u)+\epsilon)$ restores the representative RMS. Strict positivity gives $M_j>0$, and before restoration $\sum_jM_jWc_j=\sum_i m_iWh_i$ for every linear $W$: transport exactly preserves the signed population first moment. M-Rope/key-included mass changes are not assumed away and remain explicit in Prop.2 in the appendix. Prop.2 isolates the remaining RMS and compact-attention-mass effects; deployed coefficients are frozen globally and listed in the appendix.

Finally, GMC removes all unselected visual positions, leaves text states unchanged, and retains each representative's original multimodal position ID. Computation resumes from layer $p$ with $K$ visual states. At inference, GMC uses no task labels, parameter updates, auxiliary models, detectors, OCR systems, or gradients.
GMC changes only support geometry, preserving the decoder's native hidden-state interface, causal order, and original multimodal coordinates.

\section{Analysis}
\label{sec:analysis}

\subsection{From Deployed Coverage to Quantization}

\paragraph{Proposition 1 (deployed batched solver).}
$F$ is normalized, monotone, and submodular~\citep{nemhauser1978analysis}.
Let $S_t$ precede deployed batch $B_t$, let $\delta_{(j)}^t$ be the $j$th
largest remaining singleton marginal, and let
$U_t=\sum_{j=1}^{K}\delta_{(j)}^t$ and
$q_t=[F(S_t\cup B_t)-F(S_t)]/U_t$. Submodularity gives
$F(S^*)-F(S_t)\le U_t$ whenever $U_t>0$; if $U_t=0$, every remaining marginal
vanishes. Iterating the positive-gain rounds gives the deployed certificate
\begin{equation}
 \begin{aligned}
 F(S_T)&\ge c_BF(S^*)+(1-c_B)F(S_0),\\[-2pt]
 c_B&:=1-\prod_t(1-q_t).
 \end{aligned}
 \label{eq:deployed-approximation}
\end{equation}
Unlike the sequential $(1-e^{-1})$ result, $c_B$ applies to actual responsibility
batches and is logged in-pass without access to $S^*$. Frozen runs also archive
$F(S)/F(V)$, a directly measurable final-support certificate. The appendix
provides the complete batchwise proof and exact padding cases.

The appearance bank converts coverage into geometric control. For unit feature
$x_i$, let $d_i(S)=\min_{j\in S}\norm{x_i-x_j}_2$ and let $D_V(S)$ denote the
appearance facility deficit. The Gibbs client geometry yields
\begin{equation}
 \frac1N\sum_i d_i(S)^2
 \le \frac{2\tau_v}{\beta_Vp_0r_0}\,D_V(S).
 \label{eq:coverage-quantization}
\end{equation}
Here $p_0$ and $r_0$ are the minimum self-mass and selected-to-self mass ratio.
The proof uses the exact identity $r_i=e^{-d_i^2/(2\tau_v)}$ and requires no
isometry assumption. Grounded-client deficit analogously controls the best
retained carrier energy for every head query row. Appendix probe-rank, seed,
full-rank, and signed-direction interventions validate predictive carrier geometry
while separating nonnegative allocation from signed realization.

\subsection{Coverage-to-Message Error}

For fixed assignments, the weighted centroid uniquely minimizes each cluster's
pre-restoration state distortion: for
$\mathcal E_j(u)=\sum_{i\in\mathcal C_j}m_i\norm{h_i-u}^2$,
$\mathcal E_j(u)=\mathcal E_j(c_j)+M_j\norm{u-c_j}^2$. Combining this identity
with Eq.~\eqref{eq:coverage-quantization} bounds the total pre-restoration error
by four interpretable terms: facility quantization, cross-image exclusion,
spatial assignment slack, and radial-scale mismatch. RMS restoration contributes
one additional measured residual. The complete derivations are in
the appendix, importantly, no assignment or restoration error is hidden inside
the coverage term.

\paragraph{Proposition 2 (message-interface bound).}
For query/head client $g$, let $a_{gi}$ be Full visual attention,
$s_g=\sum_i a_{gi}$, and let $\mathcal N$ be the next locally
$L_{\mathcal N}$-Lipschitz RMS normalization. Weighted Cauchy-Schwarz bounds
the transported representation term by
$R_g\le\sqrt2L_{\mathcal N}s_g\sqrt{E_{\rm post}}$.
Let $M_{gj}=\sum_{i\in\mathcal C_j}a_{gi}$, let $\bar a_{gj}$ be compact
attention, and set
$H_S=\max_{j\in S}\norm{\mathcal N(\widetilde h_j)}_2$.
Carrier energy is the allocation interface, while signed realization is the
transport interface: $\{v,v\}$ and $\{v,-v\}$ have equal responsibilities yet
sums $2v$ and $0$. Adding and subtracting the transported message gives the
exact decomposition
\begin{align}
 y_g-\widetilde y_g
 &=\sum_i a_{gi}W_g^V(x_i-\widetilde x_{\pi(i)})\notag\\[-2pt]
 &\quad+\sum_j(M_{gj}-\bar a_{gj})W_g^V\widetilde x_j,\notag\\
 \norm{y_g-\widetilde y_g}_2
 &\le\norm{W_g^V}_2\left[R_g+H_S\norm{M_g-\bar a_g}_1\right].
 \label{eq:message-bound}
\end{align}
Thus coverage controls carrier quantization, transport controls the first signed
term, and compact attention-mass mismatch remains explicit; no directional
alignment or cancellation assumption is made. GMC therefore factorizes signed
compression into two auditable components: the nonnegative objective
in Eq.~\eqref{eq:coverage} allocates complementary carriers, while transport and
native compact attention realize their signed message. The appendix derives
the angular counterpart and audits exact compact-softmax message error.

Grouping Full visual logits by $\mathcal C_j$ while retaining every nonvisual key
as a singleton gives
$\norm{M_g-\bar a_g}_1\le\norm{u_g^F-u_g^S}_\infty$ by softmax's
$\ell_\infty\!\to\!\ell_1$ contraction; the appendix gives the proof.
Equation~\eqref{eq:message-bound} also identifies why selection and realization
cannot substitute for one another. Coverage can place a carrier near every required
message mode while still losing its signed population, whereas exact transport onto
poorly allocated carriers leaves a large quantization term. Their errors enter the
same bound additively, motivating the crossed interventions evaluated below.

\begin{table*}[!t]
\centering

{\small
\setlength{\tabcolsep}{0pt}

% ===========================================================================
% Qwen2.5-VL-7B
% ===========================================================================

\begin{tabular*}{\textwidth}{
@{\extracolsep{\fill}}
l l r r r r r r r r
@{}}
\toprule
Method
& Venue
& $K$
& TextVQA
& ChartQA
& MME
& POPE
& MMBench
& GQA
& Avg. \\
\midrule

\rowcolor{black!8}
\multicolumn{10}{@{}c@{}}{
\textit{Qwen2.5-VL-7B: Full-relative performance (\%)}
} \\

Full
& -
& 1296
& 100.00
& 100.00
& 100.00
& 100.00
& 100.00
& 100.00
& 100.00 \\

\rowcolor{black!8}
\multicolumn{10}{@{}c@{}}{
\textit{Retain 256 tokens ($\downarrow 80.2\%$)}
} \\

FastV
& ECCV'24
& 256
& 95.77
& 82.73
& 96.17
& 95.19
& 96.14
& 91.15
& 92.86 \\

DivPrune
& CVPR'25
& 256
& 89.07
& 79.95
& 93.11
& 97.89
& 96.62
& 98.02
& 92.44 \\

VisionZip
& CVPR'25
& 256
& 90.60
& 81.76
& 93.40
& 96.80
& 96.49
& 93.92
& 92.16 \\

CDPruner
& NeurIPS'25
& 256
& 96.83
& 84.59
& 96.68
& 93.08
& \underline{97.71}
& 97.64
& 94.42 \\

MMTok
& ICLR'26
& 256
& 90.21
& 83.80
& 96.51
& 98.56
& 95.26
& 96.05
& 93.40 \\

HAWK
& CVPR'26
& 256
& 98.35
& 89.46
& 96.68
& 97.19
& 95.60
& 90.38
& 94.61 \\

\rowcolor{black!4}
\early{}
& Ours
& 256
& \underline{99.21}
& \underline{94.67}
& \underline{97.39}
& \underline{98.57}
& 97.03
& \underline{99.81}
& \underline{97.78} \\

\rowcolor{black!9}
\faithful{}
& Ours
& 256
& \textbf{100.41}
& \textbf{102.49}
& \textbf{99.41}
& \textbf{99.97}
& \textbf{100.00}
& \textbf{99.85}
& \textbf{100.36} \\

\rowcolor{black!8}
\multicolumn{10}{@{}c@{}}{
\textit{Retain 128 tokens ($\downarrow 90.1\%$)}
} \\

FastV
& ECCV'24
& 128
& 86.72
& 62.92
& 87.34
& 88.86
& 88.04
& 83.99
& 82.98 \\

DivPrune
& CVPR'25
& 128
& 78.26
& 60.83
& 85.23
& 95.90
& 93.96
& 94.62
& 84.80 \\

VisionZip
& CVPR'25
& 128
& 82.10
& 61.50
& 86.10
& 91.60
& 90.81
& 86.76
& 83.15 \\

CDPruner
& NeurIPS'25
& 128
& 91.42
& 68.60
& 91.60
& 88.63
& 92.03
& 94.94
& 87.87 \\

MMTok
& ICLR'26
& 128
& 75.92
& 58.86
& 90.73
& 96.80
& 89.78
& 91.09
& 83.86 \\

HAWK
& CVPR'26
& 128
& 93.07
& 74.62
& 92.03
& 91.21
& 90.89
& 83.36
& 87.53 \\

\rowcolor{black!4}
\early{}
& Ours
& 128
& \underline{94.93}
& \underline{79.23}
& \underline{96.58}
& \underline{97.49}
& \underline{94.47}
& \underline{98.35}
& \underline{93.51} \\

\rowcolor{black!9}
\faithful{}
& Ours
& 128
& \textbf{98.81}
& \textbf{97.92}
& \textbf{99.01}
& \textbf{99.90}
& \textbf{99.08}
& \textbf{99.91}
& \textbf{99.11} \\

\bottomrule
\end{tabular*}

\medskip

% ===========================================================================
% LLaVA-1.5-7B
% ===========================================================================

\begin{tabular*}{\textwidth}{
@{\extracolsep{\fill}}
l l r r r r r r
@{}}
\toprule
Method
& Venue
& $K$
& MME
& POPE
& MMBench
& GQA
& Avg. \\
\midrule

\rowcolor{black!8}
\multicolumn{8}{@{}c@{}}{
\textit{
LLaVA-1.5-7B: Full-relative performance (\%);
paired values follow $K=128/64$
}
} \\

Full
& -
& 576
& 100.00/100.00
& 100.00/100.00
& 100.00/100.00
& 100.00/100.00
& 100.00/100.00 \\

FastV
& ECCV'24
& 128/64
& 80.00/67.50
& 69.40/55.90
& 86.70/74.20
& 80.10/74.50
& 79.05/68.03 \\

FitPrune
& AAAI'25
& 128/64
& 95.38/83.57
& 90.69/70.90
& 96.91/90.42
& 94.51/84.49
& 94.37/82.35 \\

VisionZip
& CVPR'25
& 128/64
& 94.60/90.80
& 96.90/89.60
& 95.80/92.90
& 93.10/89.00
& 95.10/90.58 \\

SparseVLM
& ICML'25
& 128/64
& 91.10/80.80
& 93.70/87.40
& 92.70/86.90
& 90.50/85.10
& 92.00/85.05 \\

MMTok
& ICLR'26
& 128/64
& 95.50/92.10
& \textbf{100.40}/\underline{99.90}
& 96.30/94.50
& 95.80/94.20
& 97.00/95.18 \\

ApET
& CVPR'26
& 128/64
& 96.72/92.05
& \underline{100.23}/98.25
& 93.22/90.82
& 96.14/92.70
& 96.58/93.46 \\

VLM-Pruner
& CVPR'26
& 128/64
& 94.95/94.09
& 99.42/95.69
& \underline{97.07}/\underline{95.08}
& \underline{99.18}/\underline{96.79}
& 97.66/\underline{95.41} \\

\rowcolor{black!4}
\early{}
& Ours
& 128/64
& \underline{98.82}/\underline{95.93}
& 98.24/94.76
& \underline{97.07}/93.62
& 98.13/95.76
& \underline{98.07}/95.02 \\

\rowcolor{black!9}
\faithful{}
& Ours
& 128/64
& \textbf{99.75}/\textbf{100.11}
& 99.86/\textbf{99.96}
& \textbf{99.73}/\textbf{99.73}
& \textbf{99.69}/\textbf{99.48}
& \textbf{99.76}/\textbf{99.82} \\

\bottomrule
\end{tabular*}

\medskip

% ===========================================================================
% Deployment
% ===========================================================================

\begin{tabular*}{\textwidth}{
@{\extracolsep{\fill}}
l c r r r r
@{}}
\toprule
Setting
& $N_{\rm src}/K_{\rm final}$
& Quality
& Token-layer Work
& E2E
& Prompt-KV $\downarrow$ \\
\midrule

\rowcolor{black!8}
\multicolumn{6}{@{}c@{}}{
\textit{Measured Qwen2.5-VL-7B document deployment}
} \\

\textsc{GMC-H1}
& $15{,}876/4{,}096$
& 98.87\%
& 28.45\%
& \textbf{1.258$\times$}
& \textbf{73.94\%} \\

\early{}
& $15{,}876/5{,}120$
& \textbf{98.95\%}
& 37.09\%
& 1.176$\times$
& 67.51\% \\

\bottomrule
\end{tabular*}
}

\caption{
\textbf{Performance comparisons on Qwen2.5-VL-7B and LLaVA-1.5-7B across several benchmarks.}
Values are retained performance relative to each method's corresponding Full execution (\%).
The best and second-best compressed results under each model, budget, and benchmark are shown in bold and underlined, respectively; ties receive the same formatting.
}
\label{tab:general}
\end{table*}

\begin{table*}[!t]
\small
\centering
\setlength{\tabcolsep}{1mm}

\begin{tabular*}{\linewidth}{
@{\extracolsep{\fill}}
l r r r r r r r r
@{}}
\toprule
Method
& $K$
& POPE
& AMBER
& Hall. aAcc
& Avg.
& CHAIRs$\downarrow$
& CHAIRi$\downarrow$
& Len. \\
\midrule

\rowcolor{black!8}
\multicolumn{9}{c}{
\textit{Full-token reference: 1,296 visual tokens (100\%)}
} \\

Full
& 1296
& 87.07
& 86.33
& 70.98
& 100.00
& 37.60
& 9.05
& 168.4 \\

\rowcolor{black!8}
\multicolumn{9}{c}{
\textit{Retain 256 tokens ($\downarrow 80.2\%$)}
} \\

VisionZip
& 256
& 85.78
& 84.42
& 67.61
& 97.19
& 32.60
& 8.32
& 151.5 \\

MMTok
& 256
& \underline{85.82}
& 84.42
& 67.51
& 97.15
& \underline{28.60}
& 7.77
& 154.4 \\

\rowcolor{black!4}
\early{}
& 256
& \underline{85.82}
& \underline{85.33}
& \underline{68.77}
& \underline{98.10}
& \textbf{26.80}
& \textbf{7.08}
& 147.5 \\

\rowcolor{black!9}
\faithful{}
& 256
& \textbf{87.04}
& \textbf{87.17}
& \textbf{70.03}
& \textbf{99.87}
& 28.80
& \underline{7.41}
& 153.3 \\

\rowcolor{black!8}
\multicolumn{9}{c}{
\textit{Retain 128 tokens ($\downarrow 90.1\%$)}
} \\

VisionZip
& 128
& 83.93
& 82.58
& 64.46
& 94.29
& 26.80
& 7.17
& 139.2 \\

MMTok
& 128
& 84.28
& 82.33
& 64.46
& 94.33
& 29.80
& 7.72
& 142.2 \\

\rowcolor{black!4}
\early{}
& 128
& \underline{84.88}
& \underline{83.67}
& \underline{65.51}
& \underline{95.56}
& \underline{25.80}
& \underline{6.62}
& 141.3 \\

\rowcolor{black!9}
\faithful{}
& 128
& \textbf{86.98}
& \textbf{86.75}
& \textbf{69.40}
& \textbf{99.39}
& \textbf{24.40}
& \textbf{6.08}
& 148.9 \\

\bottomrule
\end{tabular*}

\caption{
\textbf{Hallucination comparisons on Qwen2.5-VL-7B across POPE, AMBER, HallusionBench, and CHAIR.}
The best and second-best compressed results under each budget are shown in bold and underlined, respectively; lower is better for CHAIRs and CHAIRi, and ties receive the same formatting. Output length is reported without ranking.
}
\label{tab:hallucination}
\end{table*}

\subsection{Message Propagation and Visual Innovation}

At shared history $h$, let $z^F,z^S,z^0$ denote Full, compact, and zero-gated
next-token logits. On the frozen set
$C^{\rm AP}=\operatorname{Top}_k(z^F)\cup\operatorname{Top}_k(z^0)$, set
$H_C=I-|C|^{-1}\mathbf1\mathbf1^\top$ and $\nu^E=H_C(z^E-z^0)$.
The diagonal Full/null anti-prior weights use the frozen sigmoid coefficient
and temperature $\kappa_{\rm AP}=\tau_{\rm AP}=1$; the appendix gives their
elementwise definition. We define
\begin{equation}
 \operatorname{VIE}(h)=
 \frac{\norm{W_{F0}^{1/2}(\nu^F-\nu^S)}_2}
 {\max\{\norm{W_{F0}^{1/2}\nu^F}_2,\varepsilon\}},
 \label{eq:vie-definition}
\end{equation}
where the common null cancels in the numerator; it fixes the diagnostic
geometry but is absent from GMC selection.  Thus VIE measures the compression
of image-induced ranking correction rather than generic output discrepancy.

Let $e_\ell(h)$ and $\epsilon_\ell(h)$ be Full/compact text-state and message
errors at block $\ell$. Local Lipschitzness gives the recursion
$e_{\ell+1}\le a_\ell e_\ell+b_\ell\epsilon_\ell$; unrolling it bounds
$d_C(h)=\norm{H_C(z^F-z^S)}_2$ by the compacted boundary error plus all subsequent
message errors, each multiplied by its downstream amplification product. This
distinguishes cached compaction from no-cache replay and shared-history analysis
from autoregressive divergence. VIE probes the common prompt-terminal decision;
complete answers are evaluated directly rather than inferred from a one-step bound.

\paragraph{Lemma 1 (candidate-margin stability).}
For any candidates $a,b$,
$|(z_a^S-z_b^S)-(z_a^F-z_b^F)|\le\sqrt2d_C(h)$; hence a positive Full margin
$\gamma_{ab}$ cannot flip when $d_C<\gamma_{ab}/\sqrt2$. Let
$a_C^E=\arg\max_{v\in C}z_v^E$ and let
$\Gamma_C^F$ be Full's minimum represented margin. Since $a^F\in C$, for any
$\eta>0$ event inclusion gives the full-vocabulary population bound
\begin{equation}
 \begin{aligned}
 \Pr(a^S\ne a^F)\le{}&\Pr(a^S\notin C)+\Pr(d_C>\eta)\\[-2pt]
 &+\Pr(\Gamma_C^F\le\sqrt2\eta).
 \end{aligned}
 \label{eq:flip-risk}
\end{equation}
The same event bounds harmful flips on Full-correct examples. A prospective
audit freezes $C=C^{\rm AP}$ before compact logits: 0/450 conditions (0/312
unique questions) escape, including none of 27 top-1 flips. The post-hoc
winner-complete certificate and autoregressive factuality tests remain separate.

\subsection{Final Tokens and Layer Work}

Final $K$ is the community-standard quality budget, while pruning depth controls
actual decoder work. After $p$ full-length blocks,
$W(p,K)=pN+(L-p)K$ exactly counts visual token layer work and
$K_{\rm eff}=W/L$ is its full-depth equivalent. End-to-end latency additionally
captures projections, kernels, and launch costs. We report an early operating
point, \early{}, and a later high-fidelity point, \faithful{}, with both final
budget and realized execution measurements.

\section{Experiments}
\label{sec:experiments}

% Declare the full-width mechanism table before the page-7 float boundary so
% it remains in the seven-page body while retaining its later textual callout.
\begin{table*}[!t]
\small
\centering
\setlength{\tabcolsep}{1.7pt}

\begin{tabular*}{\linewidth}{
@{\extracolsep{\fill}}
l c c c r r r r r r r
@{}}
\toprule
Configuration
& Msg.
& Hist.
& Pop.
& TextVQA
& ChartQA
& MME
& POPE
& MMBench
& GQA
& Macro \\
\midrule
\rowcolor{black!8}
\multicolumn{11}{c}{
\textit{
Qwen2.5-VL-7B, L16: retain 128/1,296 visual tokens
($\downarrow 90.1\%$)
}
} \\

Current message
& \yes
& \no
& \no
& 81.39
& 76.92
& 2333
& 86.69
& 82.99
& 59.59
& 78.48 \\

+ historical views
& \yes
& \yes
& \no
& 81.23
& 75.80
& 2312
& 86.88
& 83.16
& 59.60
& 78.21 \\

+ population transport
& \yes
& \no
& \yes
& 82.22
& 76.52
& 2319
& 86.99
& 82.82
& 59.71
& 78.51 \\

+ history + population
& \yes
& \yes
& \yes
& 81.78
& 76.84
& 2319
& 87.09
& 83.16
& 59.63
& \textbf{78.55} \\

\midrule
\multicolumn{6}{l}{%
\smash{%
\rlap{%
\hspace*{-\tabcolsep}%
\raisebox{-\dp\strutbox}{%
\color{black!8}%
\rule{\linewidth}{%
\dimexpr\ht\strutbox+\dp\strutbox\relax
}%
}%
}%
}%
\textit{Factorial macro contrasts (percentage points)}
}
&
\multicolumn{5}{r}{
Effect [paired 95\% CI]
} \\

\multicolumn{6}{l}{
Historical views
}
&
\multicolumn{5}{r}{
$-0.12\;[-0.37,+0.13]$
} \\

\multicolumn{6}{l}{
Population transport
}
&
\multicolumn{5}{r}{
$\boldsymbol{+0.19\;[+0.05,+0.33]}$
} \\

\multicolumn{6}{l}{
History $\times$ Population
}
&
\multicolumn{5}{r}{
$\boldsymbol{+0.32\;[+0.05,+0.59]}$
} \\

\bottomrule
\end{tabular*}

\caption{
\textbf{History $\times$ Population factorial on Qwen2.5-VL-7B across six benchmarks.}
Configurations differ only in historical views and population transport; Macro
averages all six benchmarks with MME normalized by 2,800. Paired 95\% confidence
intervals are obtained by task-native hierarchical bootstrap; effects whose
intervals exclude zero are shown in bold.
}
\label{tab:ablation}
\end{table*}

\subsection{Setup}

We evaluate Qwen2.5-VL-7B-Instruct~\citep{bai2025qwen25vl} at $N=1{,}296$
($1008^2$) visual tokens and LLaVA-1.5-7B~\citep{liu2023llava} at
$N=576$. H2/L16 denote compaction before transformer blocks 2/16,
respectively. Fixed-token and factorial rows share the same frozen solver for
controlled comparison; document deployment bounds the number of appearance
clients and evaluates gains in chunks to support scalable inference. General
evaluation spans TextVQA~\citep{singh2019textvqa},
ChartQA~\citep{masry2022chartqa}, MME~\citep{fu2023mme},
POPE~\citep{li2023pope}, MMBench~\citep{liu2023mmbench}, and
GQA~\citep{hudson2019gqa}; faithfulness evaluation additionally includes
AMBER~\citep{wang2023amber},
HallusionBench~\citep{guan2024hallusionbench}, and
CHAIR~\citep{rohrbach2018chair}. Further details are provided in the appendix.

\subsection{General Capability at Fixed Final Tokens}

As shown in table~\ref{tab:general}, \early{} retains 93.51/97.78\% at $K=128/256$, while \faithful{} reaches
99.11/100.36\%. On LLaVA:
at $K=64/128$, \faithful{} obtains 99.82/99.76\% on the fixed common suite. The later
boundary therefore remains essentially Full-equivalent across both projectors,
whereas H2 exposes the earlier compute, fidelity operating point at identical
final representational capacity.  The independent predeclared four-task
bootstrap family confirms the Qwen $K=256$ aggregate gain after Holm
correction; complete multiple local results, confidence intervals, and the three-budget frontier are provided in
the appendix.

\subsection{Faithfulness and Hallucination}

Against VisionZip and MMTok at the identical $K=128$, improvements on POPE,
AMBER, HallusionBench, CHAIRs, and CHAIRi all survive Holm correction, CHAIR recall and paired tests make a shorter caption explanation unlikely in the appendix.
VisionZip supplies a second selector: \faithful{} improves all core accuracy and hallucination columns at both budgets. Agreement across binary, compositional, and captioning tests rejects threshold tuning, support maps
show complementary text, numeric, and relational carriers.

\begin{figure*}[!t]
\centering
\includegraphics[width=0.91\linewidth]{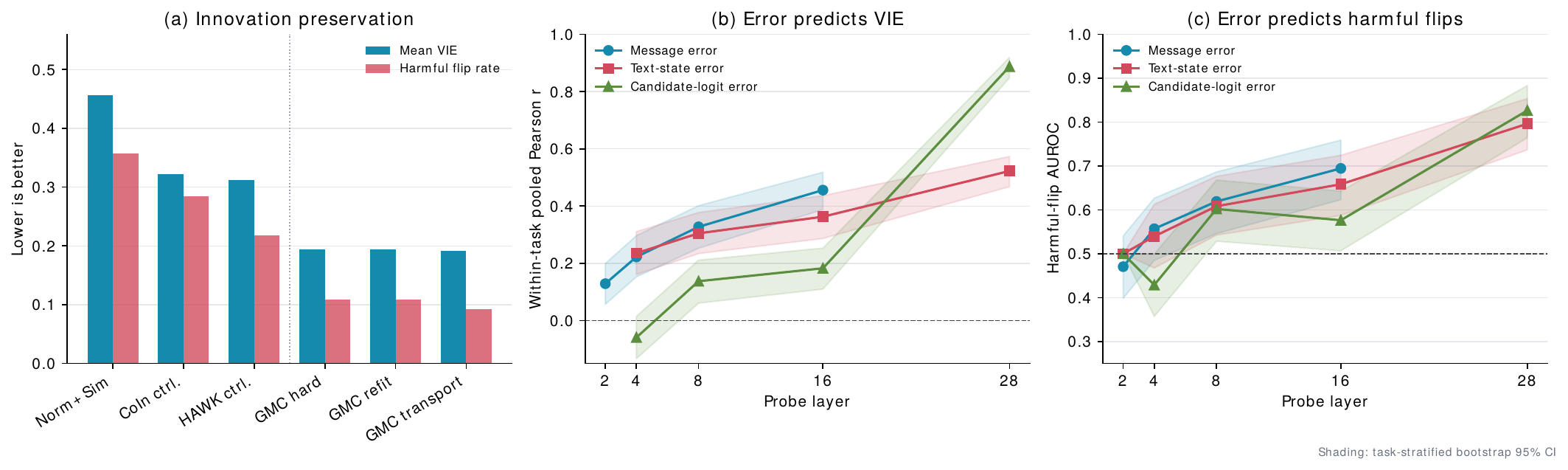}
\caption{
\textbf{Mechanism analysis of GMC-H2 on Qwen2.5-VL-7B at
$1{,}296\!\rightarrow\!256$ visual tokens.}
Same-path controls and task-stratified 95\% intervals evaluate innovation
preservation and downstream errors ($n=750$).
}
\label{fig:mechanism}
\end{figure*}

\subsection{Physical Efficiency and Architecture Transfer}

On the 15,876-token DocVQA~\citep{mathew2021docvqadatasetvqadocument}, GMC retains 98.87\% ANLS while
reaching $1.258\times$ end-to-end speedup with 73.94\% less prompt KV.
Answer-horizon and cached/no-cache audits attribute the gain to removed visual-prefix
computation, not shorter outputs or execution mismatch. Token-layer work,
prompt-KV bytes, and latency consistently verify physical compaction rather than
masking. Without architecture-specific retuning, GMC attains high
common-suite LLaVA retention at both listed budgets in
Table~\ref{tab:general}.

\subsection{Mechanism: Messages, Innovation, and Flips}
\label{sec:mechanism}

Figure~\ref{fig:mechanism} identifies support construction as the dominant
factor and candidate-logit error as the strongest predictor of harmful flips.
Rank 4-3584 and signed controls preserve L16 quality, whereas GMC leads at H2
by 3.24-6.60 points, separating carrier allocation from signed realization.
With all post-support operations fixed, GMC further exceeds the four retained
selectors at L16 by 2.74-12.61 points, rejecting context-free ranking.

With message support enabled, the History$\times$Population factorial isolates
the two realization factors. Population improves the L16 macro by $+0.19$
points, with a $+0.32$-point interaction with History. The effect of History
therefore changes from $-0.27$ points without Population to $+0.04$ with it,
while the population-enabled configurations reach 78.51 and 78.55.
Historical clients expose earlier message modes, whereas transport realizes
their discarded states on the shared support. The complete $2^3$ factorial in
the appendix likewise yields positive marginal contributions from current
messages, historical views, and population transport across complementary
settings.

% TODO before submission: inline the contents of this table into this single .tex file.
\begin{table}[!ht]
\small
\centering
\setlength{\tabcolsep}{1.2pt}
\renewcommand{\arraystretch}{0.82}
\resizebox{\linewidth}{!}{
\begin{tabular*}{\linewidth}{@{\extracolsep{\fill}}l r l r@{}}
\toprule
Support & L16 macro & Support & L16 macro \\
\midrule
\textbf{GMC} & \textbf{81.79} & Random & 74.58$^*$ \\
Uniform-head attn. & 78.93$^*$ & HAWK-style$^\S$ & 69.18$^*$ \\
Norm + div. & 76.98$^*$ & \multicolumn{2}{c}{-} \\
\bottomrule
\end{tabular*}
}
\caption{
\textbf{Same-path L16 support comparison at $K=128$.}
All later modules are fixed; $^*$ marks results significantly below GMC and
$^\S$ the HAWK-style control.
}
\label{tab:main-same-path-k128}
\end{table}

The observed distortion ratio
$\rho_x:=\sqrt{2d_C(x)}/\Gamma_C^F(x)<1$ certifies
21.6\% [18.8, 24.5] of cases with zero violations. Frozen $C^{\rm AP}$ has
0/450 prospective escapes and covers all 27 top-1 flips, auditing
Eq.~\eqref{eq:flip-risk} without an auxiliary forward pass. These audits link
coverage, signed-message error, and candidate margins to carrier availability,
compact realization, and decision stability, showing why support and transport
must be evaluated jointly.

\section{Conclusion}

We presented GMC, a training-free view of visual compression as the construction
and realization of a query-conditioned message coreset. By allocating
complementary carriers across grounded, appearance, and coordinate-aware evidence
and transporting the discarded population into those carriers, GMC preserves the
collective computation consumed by later decoder layers while enabling physical
sequence compaction. The accompanying analysis separates support coverage from
signed-message realization and connects both components to visual innovation and
candidate-margin stability; crossed interventions, architecture transfer, and
deployment audits support this mechanism across tasks, models, and compression
budgets. Together, these results establish message preservation, rather than
isolated token importance, as a foundation for faithful
VLM compression.

\nocite{yang2025visionzip,zhang2025vispruner,kim2026zooprune,
dong2026mmtok,ye2025fitprune,wang2026posprune,yu2026icctp,
xu2026visiondrop,han2026ficoco,ma2026apet,yang2025vflowopt,
choi2026docprune,zhang2025cdpruner,li2025mob,zou2025holov,
li2025btp,wang2025autoprune,wang2026informationhorizon,
gao2026quietprune}
{
    \small
    \bibliographystyle{ieeenat_fullname}
    \bibliography{main}
}

\clearpage
% \setcounter{page}{1}
% \maketitlesupplementary

% ============================================================================
% Appendix
% This file is included by the main TeX file using \input{SUP_arxiv}.
% Do not add \documentclass, \usepackage, \begin{document}, or \end{document}.
% ============================================================================

% Reset appendix counters.
\setcounter{table}{0}
\setcounter{figure}{0}
\setcounter{equation}{0}

% Appendix numbering.
\renewcommand{\thetable}{S\arabic{table}}
\renewcommand{\thefigure}{S\arabic{figure}}
\renewcommand{\theequation}{S\arabic{equation}}

\appendix

% \twocolumn[...] automatically starts a new page.
% The optional content spans both columns.
\twocolumn[
\begin{center}
  \vspace*{0.5em}
  {\LARGE\bfseries Appendix\par}
  \vspace{1.2em}
\end{center}
]

\noindent
This appendix gives the implementation details, controlled comparisons,
ablations, and proofs that complement the main paper. Unless stated otherwise,
comparisons use the same checkpoint, processor, prompt, generation
configuration, evaluator, and final physical token count.

\section{Experimental Details}
\label{sec:cond-experimental}

\subsection{Models, benchmarks, and generation}

The primary backbone is \emph{Qwen/Qwen2.5-VL-7B-Instruct}, evaluated after
bicubic $1008^2$ resizing with $N=1{,}296$ visual tokens.  GMC-H2 and GMC-L16
compact before decoder blocks 2 and 16.  The transfer backbone is
\emph{llava-hf/llava-1.5-7b-hf}, with $N=576$ and the L16 boundary.  Models are
loaded in \textsc{bfloat16}; selection accumulators use FP32; attention uses
PyTorch SDPA.  Decoding is greedy with batch size one.

The controlled MMTok row uses its released implementation with the same Qwen
checkpoint and evaluation stack.  VisionZip and FastV-style rows in the
compute comparison are locally executed mechanisms under the same inputs and
hardware.  Same-path selector experiments replace only support indices;
assignment, transport, positions, compaction, and downstream decoding remain
fixed.

The official MMTok adapter is evaluated at its released repository revision and
uses its published coverage temperatures while matching the physical target
$K$.  The VisionZip and FastV-style rows use local implementations of their
released selection mechanisms inside the same Qwen execution stack.  They are
included in fixed-$K$, equal-work, and H800 comparisons because those settings
have identical source images, processor output, prompts, and evaluator.  A Full
row always refers to the single shared uncompressed execution in
Table~\ref{tab:cond-fixed-k}; no method-specific denominator is used there.

Same-path controls comprise random support, uniform-head attention
$(HT)^{-1}\sum_{h,t}A_h[t,i]$, norm plus cosine-diversity selection, and
farthest-first feature $k$-center.
Each returns exactly $K$ source indices and then uses the common population
assignment, centroid transport, RMS restoration, original positions, physical
compaction, and downstream decoder.

\paragraph{Comparison axes.}
The experiments separate representation capacity from executed decoder work.
At fixed $K$, every method presents the same number of visual positions to the
post-compression decoder, which compares compact-prompt and prompt-KV capacity.
At fixed $W$, comparator budgets are obtained by solving
Eq.~\eqref{eq:cond-work}, which compares the number of visual token-block
interactions while allowing final support sizes to differ.  Measured latency
adds client construction, support selection, transport, compaction, vision,
and generation to this architecture-level work measure.  We report fixed-$K$
capacity, fixed-$W$ decoder work, and measured latency as complementary views
of compact execution.

The absolute-score table uses one shared Full execution for every method at a
given backbone and dataset.  Full-relative summaries divide each compressed
score by that shared Full score only after native evaluation, then average the
taskwise ratios.  For same-path interventions, all samples reuse the same
boundary, evaluator, transport, positions, and decoder; only the support
indices or recovery map named by the row changes.

\begin{table*}[t]
\centering
\small
\setlength{\tabcolsep}{2.0pt}
\begin{tabular*}{\linewidth}{@{\extracolsep{\fill}}l l r r p{0.39\linewidth}@{}}
\toprule
Task & Split & $n$ & Max new & Prompt/evaluator summary \\
\midrule
TextVQA & validation & 5,000 & 128 & Single-word-or-phrase prompt; exact-match adapter \\
ChartQA & test & 2,500 & 16 & Single-word prompt; relaxed accuracy \\
MME & test & 2,374 & 16 & Short-answer prompt; category-pair total \\
POPE & test & 9,000 & 128 & Short-answer prompt; accuracy/F1 evaluation \\
AMBER & test & 1,200 & 128 & Short-answer prompt; generative hallucination metrics \\
HallusionBench & image & 951 & 8 & Yes/no-only prompt; paired-set evaluator \\
CHAIR & validation & 500 & 512 & Detailed caption prompt; official OPERA scorer \\
MMBench-EN & test & 4,329 & 8 & Lettered options; option-letter evaluator \\
DocVQA & validation & 5,349 & 32 & Single-word-or-phrase prompt; ANLS \\
\bottomrule
\end{tabular*}
\caption{Evaluation tasks and generation limits.  Qwen, LLaVA, Full, and all
compressed variants use their corresponding shared checkpoint, processor,
prompt, generation rule, and evaluator.}
\label{tab:cond-eval}
\end{table*}

\begin{table*}[t]
\centering
\small
\setlength{\tabcolsep}{2.2pt}
\begin{tabular*}{\linewidth}{@{\extracolsep{\fill}}r l r r r r r r@{}}
\toprule
$K$ & Method & TextVQA & ChartQA & MME & POPE & Macro & Work \\
\midrule
\rowcolor{black!8}128 & Full & 82.64 & 78.76 & 2340.4 & 87.07 & 100.00 & 100.00 \\
128 & MMTok & 62.74 & 46.36 & 2123.5 & 84.28 & 80.58 & 9.88 \\
128 & VisionZip & 62.82 & 56.24 & 2190.9 & 84.06 & 84.39 & 9.88 \\
128 & FastV-style H2 & 72.31 & 36.24 & 2080.6 & 79.93 & 78.55 & 16.31 \\
128 & \early & 78.45 & 62.40 & 2260.3 & 84.88 & 92.06 & 16.31 \\
128 & \faithful & 81.66 & 77.12 & 2317.1 & 86.98 & \textbf{98.91} & 61.38 \\
\midrule
\rowcolor{black!8}256 & Full & 82.64 & 78.76 & 2340.4 & 87.07 & 100.00 & 100.00 \\
256 & MMTok & 74.55 & 66.00 & 2258.6 & 85.82 & 92.27 & 19.75 \\
256 & VisionZip & 73.97 & 70.28 & 2311.5 & 86.38 & 94.18 & 19.75 \\
256 & FastV-style H2 & 79.32 & 59.56 & 2243.0 & 84.68 & 91.18 & 25.49 \\
256 & \early & 81.99 & 74.56 & 2279.2 & 85.82 & 97.46 & 25.49 \\
256 & \faithful & 82.98 & 80.72 & 2326.6 & 87.04 & \textbf{100.57} & 65.61 \\
\midrule
\rowcolor{black!8}512 & Full & 82.64 & 78.76 & 2340.4 & 87.07 & 100.00 & 100.00 \\
512 & MMTok & 79.26 & 73.56 & 2279.7 & 86.41 & 96.49 & 39.51 \\
512 & VisionZip & 80.49 & 77.52 & 2354.2 & 87.00 & 99.08 & 39.51 \\
512 & FastV-style H2 & 83.15 & 77.08 & 2358.5 & 86.79 & 99.74 & 43.83 \\
512 & \early & 83.07 & 80.56 & 2327.1 & 86.40 & 100.37 & 43.83 \\
512 & \faithful & 83.66 & 81.40 & 2337.3 & 86.99 & \textbf{101.09} & 74.07 \\
\bottomrule
\end{tabular*}
\caption{Absolute scores under one Qwen2.5-VL protocol.  Macro is the
unweighted mean of taskwise retention relative to the shared Full row.  Work
is visual token-layer retention (\%), not final-token retention.}
\label{tab:cond-fixed-k}
\end{table*}

\subsection{Statistics and timing}
\label{sec:cond-statistics}

Quality intervals use paired bootstrap resampling at each benchmark's natural
independent unit; image-linked records move together.  Equal-task macros first
normalize MME by its attainable total and then average task scores.  H800
system comparisons share SDPA, batch one, synchronized event timing, and the
same examples.  The cached document deployment uses an A800, matched timing
offsets, and 16 generated tokens.  We report final $K$, visual token-layer
work, prompt-KV reduction, elapsed time, and peak allocation separately.

For task scores, a bootstrap draw resamples the same example indices for Full,
GMC, and the comparator, preserving pairing.  TextVQA, ChartQA, POPE, AMBER,
and DocVQA resample questions or images as defined by their evaluator;
HallusionBench moves official paired-set clusters together; MME resamples its
category-consistent units before reconstructing the category total.  Reported
confidence intervals are percentile intervals from 20,000 paired replicates.
The factorial tables use the same paired unit and compute each contrast inside
every replicate.

The equal-task macro prevents large datasets from dominating a pooled count.
For Qwen's four-task tables it averages TextVQA, ChartQA, POPE, and MME after
mapping MME to percentage of its attainable category total.  Full-relative
retention is used only when the table explicitly labels it; absolute-score
tables keep every benchmark in its native metric.

Timing uses synchronized GPU execution after warm-up and reports the full path
from processed model inputs through generation.  Prompt-KV reduction is
computed from the actual compact sequence and number of cached layers, while
peak allocation is measured over the whole run.  Token-layer work in
Eq.~\eqref{eq:cond-work} is an architecture-level compute proxy; wall-clock
measurements separately include vision encoding, client construction,
selection, transport, compaction, decoder kernels, and generation.

Let $u=1,\ldots,n$ index the benchmark's independent sampling units and let
$g_m(u)$ be the decomposable score contribution of method $m$.  A paired
bootstrap replicate draws indices $u_1^*,\ldots,u_n^*$ once and evaluates all
methods on that same multiset.  For a comparator $c$, each replicate records
\begin{equation}
 \Delta^*=\mathcal E(\{g_m(u_j^*)\}_{j=1}^n)
 -\mathcal E(\{g_c(u_j^*)\}_{j=1}^n),
 \label{eq:cond-bootstrap}
\end{equation}
where $\mathcal E$ is the task's native evaluator or the equal-task macro.
Percentile 2.5 and 97.5 quantiles form the reported interval.  For MME and
HallusionBench, the resampled unit contains every linked record needed by the
official aggregate, so a replicate never breaks an evaluator-defined group.

The two confirmatory multiplicity families use Holm's step-down procedure.
Within a family of $m$ endpoints, raw paired-bootstrap $p$-values are sorted as
$p_{(1)}\le\cdots\le p_{(m)}$ and compared with
$\alpha/(m-j+1)$ in order.  The general family contains the three L16
fixed-budget macro comparisons with Full; the faithfulness family contains the
five low-budget hallucination endpoints.  Descriptive ablations and mechanism
correlations retain their paired intervals and are not pooled into either
family.

Elapsed time is averaged after one untimed warm-up pass and explicit device
synchronization at every interval boundary.  The same processed examples and
generation limit are used for Full and compressed rows.  Peak allocation is
reset immediately before the measured path, while prompt-KV bytes are computed
from sequence length, cached layers, heads, head dimension, and element size.
Thus peak allocation includes temporary buffers and model-resident memory,
whereas KV reduction isolates the persistent prompt state affected by visual
compaction.

\section{Controlled Quality and Compute Comparisons}

\subsection{Fixed final-token capacity}

At equal compact-prompt capacity, L16 provides the highest-fidelity GMC point,
whereas H2 uses substantially less decoder work.  At $K=128$, the paired
L16-minus-MMTok differences are $+18.91$ [17.71,20.16] on TextVQA,
$+30.76$ [28.72,32.84] on ChartQA, and $+2.70$ [2.22,3.18] on POPE.

Table~\ref{tab:cond-fixed-k} uses one Full denominator and native absolute task
scores, so its comparisons do not depend on the baseline reported by a source
paper.  The two GMC boundaries expose the role of dense-prefix depth.  H2
processes only two blocks before compaction and retains 16.31\% of visual
token-layer work at $K=128$; L16 retains 61.38\% because its compact prompt is
formed after deeper visual-text mixing.  Final $K$ is therefore a common
representation capacity, while Work records the distinct amount of dense
decoder computation used to construct it.

The quality gap is concentrated in evidence-sensitive tasks.  At $K=128$, H2
already retains substantially more TextVQA and ChartQA evidence than the other
early selectors, while L16 approaches the shared Full scores across all four
tasks.  Increasing $K$ reduces the need to represent multiple local evidence
regions with one carrier.  At $K=256$, L16 matches Full overall and obtains a
small aggregate gain, driven principally by ChartQA; at $K=512$, both GMC
boundaries operate close to the Full task average.

The Full-relative Macro and native task columns serve complementary purposes.
Macro puts heterogeneous metrics on a common retention scale, while the native
columns reveal where that aggregate originates.

\subsection{Paired fidelity and faithfulness intervals}

Paired uncertainty follows the same fixed-$K$ trend.  Relative to the shared
Full execution, GMC-L16 macro retention is 98.91\% [98.28,99.53] at $K=128$,
100.57\% [100.00,101.14] at $K=256$, and 101.09\%
[100.61,101.58] at $K=512$.  The smallest marginal taskwise retention implied
by a 95\% lower confidence bound is 96.14\%, 97.78\%, and 98.52\%,
respectively.  MME has the widest interval because inference resamples its
1,187 category-stratified image pairs rather than treating the paired questions
as independent.

The hallucination evaluations use the same final token budgets but cover three
decision formats.  At $K=128$, GMC-L16 minus controlled MMTok is
$+2.70$ [2.22,3.18] on POPE accuracy, $+4.42$ [2.95,5.93] on AMBER accuracy,
and $+4.94$ [2.24,7.85] on HallusionBench aAcc.  In open captioning, lower is
better: CHAIRs changes by $-5.40$ [$-9.80,-1.20$] and CHAIRi by
$-1.65$ [$-2.86,-0.44$].  The classification and caption metrics therefore
move in the same favorable direction under the severe bottleneck.

Two multiplicity families are evaluated separately.  The general family tests
the three GMC-L16 macro retentions against Full; after Holm correction,
$K=256$ and $K=512$ remain resolved.  The faithfulness family contains the
five $K=128$ endpoints above, and all five remain resolved after correction.
These family-wise results summarize joint uncertainty without replacing the
native task metrics in Table~\ref{tab:cond-fixed-k}.

\subsection{Equal token-layer work}

\begin{table*}[t]
\centering
\small
\setlength{\tabcolsep}{1.8pt}
\begin{tabular*}{\linewidth}{@{\extracolsep{\fill}}l l c c c c c@{}}
\toprule
Target & Method & $K$ & \shortstack{Final retention(\%)} & Work & Macro & $\Delta$ comparator$-$GMC [95\% CI] \\
\midrule
\rowcolor{black!8}H2-128 & \early & 128 & 9.88 & 16.31 & 92.06 & - \\
H2-128 & MMTok & 211 & 16.28 & 16.28 & 89.66 & $-2.40$ [$-3.38,-1.43$] \\
H2-128 & VisionZip & 211 & 16.28 & 16.28 & 91.82 & $-0.24$ [$-1.15,+0.69$] \\
H2-128 & FastV-style H2 & 128 & 9.88 & 16.31 & 78.55 & $-13.50$ [$-14.55,-12.45$] \\
\midrule
\rowcolor{black!8}H2-256 & \early & 256 & 19.75 & 25.49 & 97.46 & - \\
H2-256 & MMTok & 330 & 25.46 & 25.46 & 94.59 & $-2.87$ [$-3.62,-2.11$] \\
H2-256 & VisionZip & 330 & 25.46 & 25.46 & 96.24 & $-1.22$ [$-1.96,-0.45$] \\
H2-256 & FastV-style H2 & 256 & 19.75 & 25.49 & 91.18 & $-6.29$ [$-7.23,-5.35$] \\
\midrule
\rowcolor{black!8}L16-128 & \faithful & 128 & 9.88 & 61.38 & 98.91 & - \\
L16-128 & MMTok & 795 & 61.34 & 61.34 & 98.20 & $-0.71$ [$-1.42,+0.02$] \\
L16-128 & VisionZip & 795 & 61.34 & 61.34 & 99.81 & $+0.90$ [$+0.27,+1.54$] \\
L16-128 & FastV-style H2 & 757 & 58.41 & 61.38 & 101.09 & $+2.18$ [$+1.63,+2.75$] \\
\midrule
\rowcolor{black!8}L16-256 & \faithful & 256 & 19.75 & 65.61 & 100.57 & - \\
L16-256 & MMTok & 850 & 65.59 & 65.59 & 98.66 & $-1.91$ [$-2.58,-1.21$] \\
L16-256 & VisionZip & 850 & 65.59 & 65.59 & 99.73 & $-0.84$ [$-1.42,-0.25$] \\
L16-256 & FastV-style H2 & 816 & 62.96 & 65.61 & 101.16 & $+0.59$ [$+0.10,+1.10$] \\
\bottomrule
\end{tabular*}
\caption{Equal visual token-layer work.  Comparator budgets solve
$28K_{\rm input}=W(p,K_{\rm GMC})$.  Intervals are paired four-task macro
differences.  Final retention is $100K/1{,}296$; Work is retained visual
token-layer computation.}
\label{tab:cond-equal-work}
\end{table*}

\begin{figure*}[t]
\centering
\includegraphics[width=0.76\linewidth]{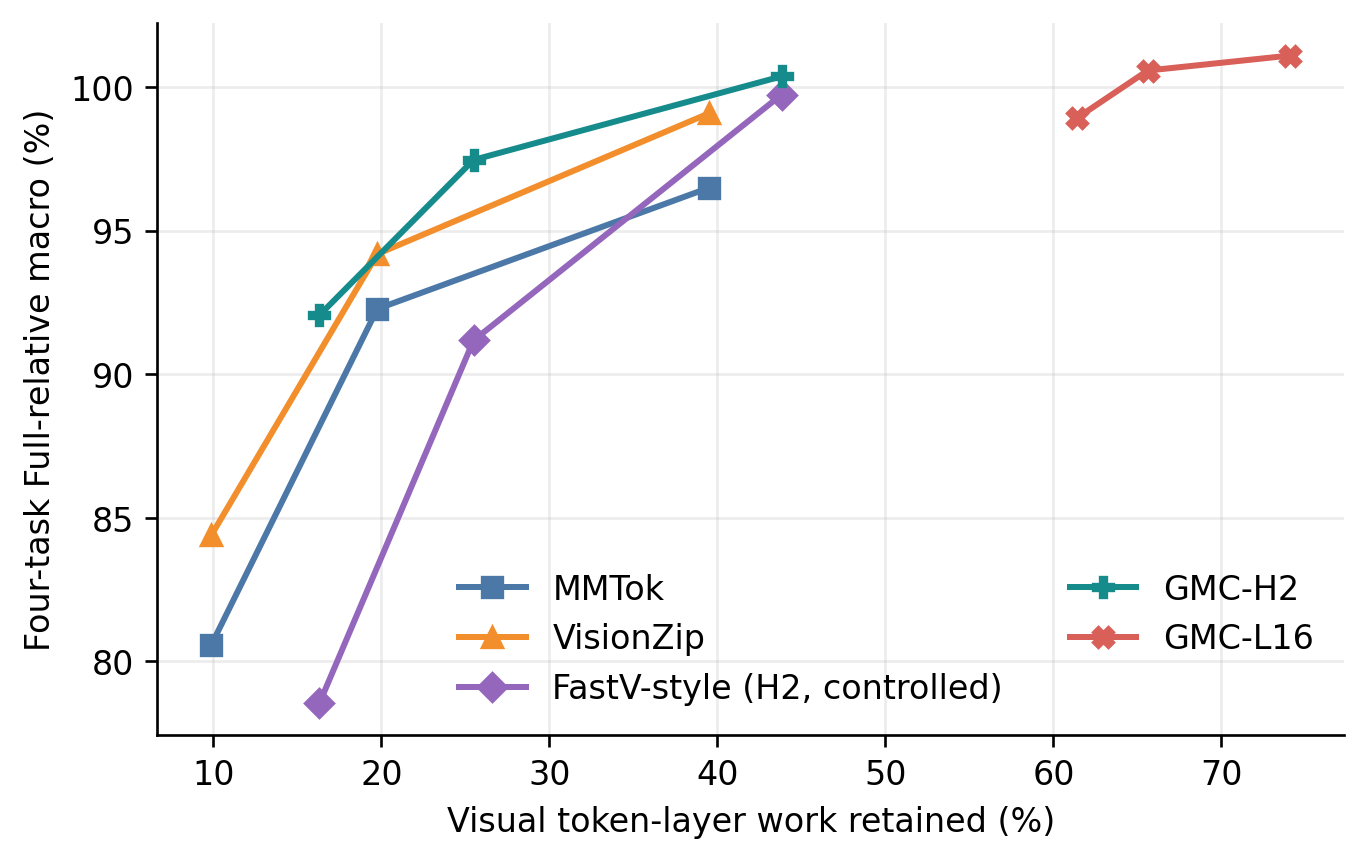}
\caption{Four-task quality versus visual token-layer work.  H2 forms the
compute-oriented GMC frontier; L16 spends more dense-prefix work to produce a
higher-fidelity compact prompt.}
\label{fig:cond-quality-work}
\end{figure*}

H2 remains competitive when comparator budgets are increased to match its
decoder work, leading MMTok at both displayed budgets and VisionZip at
$K=256$, while ending with only 9.88\% or 19.75\% of the source visual tokens.
L16 uses the same compact final supports as H2 to provide a high-fidelity
operating point.

The comparator budgets in Table~\ref{tab:cond-equal-work} solve the work
equation directly.  An input-boundary method receives $K=211$ tokens to match
H2-128 and $K=330$ to match H2-256.  Under these larger supports, H2 retains a
$2.40$-point macro advantage over MMTok at the first target and a
$2.87$-point advantage at the second.  VisionZip is statistically tied at the
narrower target and trails by $1.22$ points at H2-256.  These comparisons
show that H2 combines competitive equal-work quality with smaller final
retention: 9.88\% versus 16.28\% at the first target and 19.75\% versus
25.46\% at the second.

At L16-equivalent work, input-side comparators require 58.41-61.34\% final
retention for the 128-token target and 62.96-65.59\% for the 256-token target.
L16 reaches its high-fidelity operating points with only 9.88\% and 19.75\%
final retention, respectively, producing a substantially smaller prompt and
prompt-KV representation.  The quality-work curve in
Figure~\ref{fig:cond-quality-work} displays both GMC choices: H2 traces the
lower-work frontier, while L16 occupies the high-fidelity compact-prompt
region.

\section{Support Allocation and Population Realization}

\subsection{Same-path selector comparison}

\begin{table*}[t]
\centering
\small
\setlength{\tabcolsep}{2.2pt}
\begin{tabular*}{\linewidth}{@{\extracolsep{\fill}}l c c c c@{}}
\toprule
Selector & H2 macro & $\Delta$ vs GMC [95\% CI] & L16 macro & $\Delta$ vs GMC [95\% CI] \\
\midrule
Random support (3 seeds) & 66.46 & $-8.65$ [$-9.27,-8.04$] & 74.58 & $-7.21$ [$-7.74,-6.68$] \\
Uniform-head attention & 67.72 & $-7.39$ [$-8.14,-6.64$] & 78.93 & $-2.86$ [$-3.42,-2.32$] \\
Norm + diversity & 70.55 & $-4.56$ [$-5.27,-3.87$] & 76.98 & $-4.81$ [$-5.42,-4.19$] \\
Feature $k$-center & 73.22 & $-1.90$ [$-2.55,-1.23$] & 79.05 & $-2.74$ [$-3.30,-2.18$] \\
\rowcolor{black!8}GMC message support & \textbf{75.11} & - & \textbf{81.79} & - \\
\bottomrule
\end{tabular*}
\caption{Same-path support intervention at $K=128$.  Only support indices
change; assignment, transport, original positions, physical compaction, and
downstream decoding are shared.  Macros average TextVQA, ChartQA, and POPE.}
\label{tab:cond-selector}
\end{table*}

Message support has a clear independent advantage at H2 and L16, exceeding
random, independent attention, norm-diversity, and feature-coverage selectors
under the same assignment, transport, positions, and downstream execution.

Random support remains substantially below GMC at both boundaries even though
it receives the same transport and position-preserving execution.  The
$7.21$-point L16 gap shows that late-layer token diffusion does not make the
choice of representatives arbitrary.  Uniform-head attention improves over
random at L16 but leaves complementary regions uncovered, while norm-diversity
and feature $k$-center test increasingly stronger forms of generic visual
coverage.  GMC's advantage over $k$-center at both boundaries identifies
query-conditioned message geometry beyond boundary-feature dispersion.

The next factorial separates this support advantage from recovery: support
determines which evidence carriers remain, while the realized compact
representation also depends on how discarded states are assigned and
transported.

\subsection{Selector $\times$ recovery}

\begin{table*}[t]
\centering
\small
\setlength{\tabcolsep}{2.4pt}
\begin{tabular*}{\linewidth}{@{\extracolsep{\fill}}l c c c c c c@{}}
\toprule
& \multicolumn{3}{c}{H2 macro} & \multicolumn{3}{c}{L16 macro} \\
\cmidrule(lr){2-4}\cmidrule(lr){5-7}
Support selector & Hard & Uniform & Grounded & Hard & Uniform & Grounded \\
\midrule
\rowcolor{black!8}GMC responsibility & 57.67 & 59.04 & \textbf{61.05} & \textbf{80.55} & 78.41 & 79.90 \\
Random & 40.51 & 46.84 & \textbf{48.01} & 49.28 & 60.29 & \textbf{65.03} \\
Norm + diversity & 43.65 & 50.22 & \textbf{51.47} & 46.05 & 58.07 & \textbf{70.82} \\
Feature $k$-center & 53.88 & 54.26 & \textbf{57.31} & 57.44 & 66.63 & \textbf{74.10} \\
\bottomrule
\end{tabular*}
\caption{Selector-recovery factorial at $1{,}296\!\rightarrow\!128$ over
TextVQA, ChartQA, and DocVQA subsets.  Hard deletion, uniform centroid merging,
and grounded population transport share the same support and execution path.}
\label{tab:cond-selector-recovery}
\end{table*}

At H2, grounded transport improves GMC over hard deletion by $+3.37$
[1.17,5.55].  At L16, the already strong GMC support is stable across recovery
maps, while grounded transport substantially improves incomplete random,
norm-diversity, and $k$-center supports.  Support allocation chooses evidence
carriers; transport determines how discarded population information is
realized on those carriers.

The full factorial separates these interfaces without changing the pruning
boundary or final token count.  Under hard deletion at H2, GMC support exceeds
random by 17.16 macro points and feature $k$-center by 3.79 points.  Grounded
transport raises all four H2 supports, with the largest absolute need appearing
when the selected set incompletely covers the population.  This behavior is
expected from weighted centroid recovery: a representative can summarize
discarded neighbors only if its cluster contains coherent evidence.

At L16, generic supports benefit strongly from population realization.  Random
support rises from 49.28 under deletion to 65.03 under grounded transport, and
norm-diversity rises from 46.05 to 70.82.  GMC's hard support is already
strong because deeper contextual states carry more integrated information;
the three GMC recovery columns consequently lie in a narrower quality regime.
The two interfaces are therefore complementary rather than interchangeable:
better recovery cannot make an arbitrary support equal to a grounded one, and
better support reduces the burden placed on each transported representative.

\subsection{Message-pathway effects}

\begin{table*}[t]
\centering
\small
\setlength{\tabcolsep}{1.9pt}
\begin{tabular*}{\linewidth}{@{\extracolsep{\fill}}l l l l l@{}}
\toprule
\multicolumn{5}{c}{\textit{(a) Balanced $2^3$ factorial effects at $K=128$}} \\
Boundary & Current message & Historical views & Population transport & Three-way interaction \\
\midrule
H2 & $+0.47$ [$+0.07,+0.88$] & $+0.53$ [$+0.15,+0.92$] & $+0.56$ [$+0.35,+0.78$] & $+0.32$ [$-0.49,+1.15$] \\
L16 & $+1.62$ [$+1.33,+1.90$] & $+0.64$ [$+0.35,+0.94$] & $+0.74$ [$+0.57,+0.92$] & $+1.45$ [$+0.87,+2.03$] \\
\midrule
\multicolumn{5}{c}{\textit{(b) Complete L16 message path minus no-message path}} \\
$K$ & TextVQA & ChartQA & POPE & Equal-task macro \\
\midrule
128 & $+5.95$ [$+5.05,+6.85$] & $+4.88$ [$+3.32,+6.44$] & $+1.77$ [$+1.44,+2.10$] & $+3.36$ [$+2.81,+3.92$] \\
256 & $+1.76$ [$+1.16,+2.37$] & $+1.00$ [$-0.08,+2.08$] & $+0.56$ [$+0.32,+0.79$] & $+0.90$ [$+0.48,+1.32$] \\
512 & $+0.45$ [$+0.02,+0.89$] & $+0.28$ [$-0.52,+1.08$] & $+0.34$ [$+0.18,+0.52$] & $+0.50$ [$+0.13,+0.89$] \\
\midrule
\multicolumn{5}{c}{\textit{(c) Sparse-reading pathway effects on DocVQA ANLS}} \\
Backbone/budget & Effect & 95\% CI & \multicolumn{2}{l}{Paired rate interaction} \\
\midrule
Qwen, $K=128$ & $+23.12$ & [$+22.02,+24.24$] & \multicolumn{2}{l}{$\Delta_{128}-\Delta_{256}=+12.93$ [$+11.79,+14.07$]} \\
Qwen, $K=256$ & $+10.19$ & [$+9.40,+11.00$] & \multicolumn{2}{l}{-} \\
LLaVA, $K=64$ & $+4.08$ & [$+3.41,+4.75$] & \multicolumn{2}{l}{-} \\
\bottomrule
\end{tabular*}
\caption{Message-pathway evidence.  Factorial effects average over all cells
of the other factors.  The rate table holds appearance/spatial banks,
assignment, residual transport, positions, and budget fixed while toggling
message clients, history, and message-derived transport weights.}
\label{tab:cond-message-path}
\end{table*}

Current messages, historical views, and population weighting each contribute
at both boundaries.  The complete pathway matters most under scarce support:
its L16 macro effect falls smoothly from $+3.36$ at $K=128$ to $+0.50$ at
$K=512$.  The H2 macro gain is $+1.65$ at $K=128$ and similarly contracts as
the budget widens.  DocVQA provides the strongest sparse-reading test, with a
resolved $+12.93$-point low-budget interaction on Qwen and an independent
LLaVA gain.

The balanced $2^3$ design toggles current-message clients, historical views,
and message-derived population weighting while averaging each main effect over
the other four cells.  At H2 all three marginal intervals exclude zero.  L16
shows a larger current-message effect and a positive three-way interaction,
indicating that history and population weighting are most useful when current
message clients also allocate the support.  The interaction is measured on the
same examples and does not rely on comparing unrelated ablation rows.

The rate experiment holds appearance and spatial banks, assignment, residual
transport machinery, positions, boundary, and evaluator fixed.  Only the
complete message pathway is enabled or removed.  Its contribution decreases
monotonically as the support widens on TextVQA and in the equal-task macro.
This pattern matches the coreset interpretation: message responsibilities are
most valuable when one representative must account for several answer-bearing
regions, and broader supports increasingly recover those regions through
appearance coverage alone.

DocVQA strengthens this interpretation because ANLS depends on sparse local
text rather than broad scene recognition.  The $+23.12$ effect at 128 tokens
contracts to $+10.19$ at 256 tokens on the same 5,349 Qwen questions, producing
the paired $+12.93$ rate interaction.  The LLaVA-1.5 pathway yields a positive
$+4.08$ effect at 64 tokens.  The repeated
low-budget effect across tokenizers, visual encoders, and decoder architectures
links the message pathway to distributed evidence preservation.

\section{Runtime and Deployment}
\label{sec:cond-runtime}

\begin{table*}[t]
\centering
\small
\setlength{\tabcolsep}{1.6pt}
\begin{tabular*}{\linewidth}{@{\extracolsep{\fill}}r l r r r r r r@{}}
\toprule
\multicolumn{8}{c}{\textit{(a) H800, standard prefix $N=1{,}296$, 300 DocVQA examples, 64 generated tokens}} \\
$K$ & Method & ANLS & Work & KV red. & Time (s) & Speedup & Peak GiB \\
\midrule
128 & Full & .9461 & 100.00 & 0.00 & .3331 & 1.000 & 15.99 \\
128 & MMTok & .4246 & 9.88 & 87.26 & .3012 & 1.106 & 15.74 \\
128 & VisionZip & .3651 & 9.88 & 87.26 & .6150 & .542 & 19.40 \\
128 & FastV-style H2 & .3382 & 16.31 & 87.26 & .2908 & 1.146 & 15.76 \\
128 & \early & .5219 & 16.31 & 87.26 & .3231 & 1.031 & 15.78 \\
128 & \faithful & .7824 & 61.38 & 87.26 & .3357 & .992 & 15.82 \\
\addlinespace[1pt]
256 & Full & .9461 & 100.00 & 0.00 & .3317 & 1.000 & 15.99 \\
256 & MMTok & .6671 & 19.75 & 77.69 & .3140 & 1.056 & 15.74 \\
256 & VisionZip & .5481 & 19.75 & 77.69 & .6112 & .543 & 19.40 \\
256 & FastV-style H2 & .4575 & 25.49 & 77.69 & .3065 & 1.082 & 15.76 \\
256 & \early & .6631 & 25.49 & 77.69 & .3299 & 1.006 & 15.78 \\
256 & \faithful & .8977 & 65.61 & 77.69 & .3411 & .972 & 15.82 \\
\midrule
\multicolumn{8}{c}{\textit{(b) H800, long prefix $N=15{,}876$, $K=4{,}096$, 16 generated tokens}} \\
$K$ & Method & ANLS & Work & KV red. & Time (s) & Speedup & Peak GiB \\
\midrule
4096 & Full & .9436 & 100.00 & 0.00 & 10.5639 & 1.000 & 28.55 \\
4096 & MMTok & .9302 & 25.80 & 74.00 & 18.0976 & .584 & 28.55 \\
4096 & FastV-style H2 & .9294 & 31.10 & 74.00 & 10.9922 & .961 & 28.49 \\
4096 & GMC-H1 & \textbf{.9619} & 28.45 & 74.00 & 10.1526 & \textbf{1.041} & 28.46 \\
4096 & \early & .9517 & 31.10 & 74.00 & 10.3085 & 1.025 & 28.47 \\
\bottomrule
\end{tabular*}
\caption{Measured quality, work, prompt-KV, latency, and peak allocation on
one H800/SDPA path.  Standard inputs expose fixed overhead; the long prefix
amortizes support construction over substantially more decoder work.}
\label{tab:cond-runtime}
\end{table*}

\begin{table*}[t]
\centering
\small
\setlength{\tabcolsep}{2.0pt}
\begin{tabular*}{\linewidth}{@{\extracolsep{\fill}}l r r r r r r@{}}
\toprule
\multicolumn{7}{c}{\textit{(a) GMC compression overhead (ms), synchronized H800 timing}} \\
$N$/$K$ & Message & Selection & Transport & Compaction & Total & Context \\
\midrule
1,296/128 & 3.9 & 38.6 & 4.7 & 17.3 & 64.5 & standard \\
1,296/256 & 4.0 & 41.4 & 4.9 & 15.4 & 65.7 & standard \\
15,876/4,096 & 18.0 & 68.5 & 21.1 & 1.3 & 108.9 & long prefix \\
\midrule
\multicolumn{7}{c}{\textit{(b) Cached A800 document deployment, $N=15{,}876$}} \\
Boundary/$K$ & ANLS & Full$-$GMC [95\% CI] & Select & Transport & E2E speedup & KV red. \\
\midrule
H1/4096 & .9326 & 1.06 [$-0.86,+3.07$] & 69.10 ms & 27.02 ms & 1.258$\times$ & 73.94\% \\
H2/5120 & \textbf{.9333} & 0.99 [$-0.87,+3.02$] & 114.33 ms & 83.64 ms & 1.176$\times$ & 67.51\% \\
\bottomrule
\end{tabular*}
\caption{Compression-side cost and the cached document operating points.
End-to-end cached time includes vision, dense prefix blocks, selection,
transport, compaction, remaining prefill, and 16 generated tokens.}
\label{tab:cond-overhead-cached}
\end{table*}

At standard resolution, vision and short generation dominate the roughly
0.33-second Full path, so compression primarily reduces prompt-KV capacity.
At $N=15{,}876$, GMC-H1 and H2 improve the same-SDPA quality-time balance.
The cached and no-cache GMC executions agree after answer normalization on
approximately 98\% of the paired examples.

The standard-prefix panel measures the regime used by the controlled quality
tables.  All compressed methods reduce prompt-KV by the same amount at a fixed
$K$, but their latency differs because support construction and execution
paths differ.  GMC-H2 remains within 3.1\% of Full time at $K=128$ and within
0.6\% at $K=256$ while delivering markedly higher ANLS than the other early
selectors.  L16 spends more dense-prefix work and therefore operates near Full
latency.  This is consistent with its role as the high-fidelity compact-prompt
point rather than the minimum-work point.

The component timing separates algorithm cost from decoder savings.  At
$N=1{,}296$, selection contributes roughly 39-41 ms and dominates the
64-66 ms compression stage; message construction and transport together
require under 10 ms.  Physical compaction costs another 15-17 ms.  These
mostly fixed costs explain why a short 0.33-second workload obtains substantial
KV reduction but little end-to-end acceleration.

At $N=15{,}876$, the bounded appearance bank keeps message construction,
selection, and transport to 108.9 ms in total.  The dense visual prefix is now
large enough for later-layer savings to amortize that cost.  GMC-H1 obtains
ANLS .9619 and a 1.041$\times$ speedup in the common no-cache H800 path; H2
obtains .9517 and 1.025$\times$.  MMTok and FastV-style use less or comparable
token-layer work but do not improve this quality-time point under the same
SDPA execution.

Prompt-KV reduction and peak allocation measure different resources.  KV
tracks the memory that scales with prompt length, retained token count, and
cached layers; whole-run peak allocation also includes model weights, vision
activations, temporary client buffers, and allocator reserve.  The long-prefix
rows reduce prompt-KV by 74\% while whole-run peaks remain close because model
weights and dense prefix activations dominate the absolute maximum.

The cached A800 path gathers dense prefix-layer prompt K/V at selected indices
and inserts transported states from the compression boundary forward.  It
therefore avoids replaying the dense prompt during generation.  H1 reaches a
1.258$\times$ end-to-end speedup with 73.94\% KV reduction; H2 reaches
1.176$\times$ with 67.51\% KV reduction.  Their paired ANLS intervals relative
to Full include zero.  The approximately 98\% normalized-answer agreement with
the no-cache execution indicates that the cache implementation preserves the
intended compact prompt behavior on nearly all evaluated examples.

\section{Cross-Architecture Transfer}

\begin{table*}[t]
\centering
\small
\setlength{\tabcolsep}{1.4pt}
\begin{tabular*}{\linewidth}{@{\extracolsep{\fill}}l r r r r r r r r@{}}
\toprule
Method & $K$ & TextVQA & MME & POPE & MMBench & AMBER & Hallusion & Mean ret. \\
\midrule
\rowcolor{black!8}Full & 576 & 48.60 & 1817.47 & 85.46 & 64.60 & 78.58 & 51.95 & 100.00 \\
\faithful & 128 & 47.94 & 1812.89 & 85.34 & 64.78 & 78.75 & 51.31 & 99.59 \\
\faithful & 64 & 47.57 & \textbf{1819.39} & 85.43 & 64.60 & \textbf{78.83} & 51.84 & 99.68 \\
\bottomrule
\end{tabular*}
\caption{LLaVA-1.5-7B transfer with 576 source tokens.  Mean retention is the
arithmetic mean of six taskwise Full-score ratios.  At $K=64$, enabling the
complete message pathway improves TextVQA by $+4.00$ [3.17,4.84] and the
equal-task macro by $+1.16$ [0.69,1.62].}
\label{tab:cond-llava}
\end{table*}

The same training-free construction transfers without architecture-specific
retuning.  The narrowest 64-token point retains 99.68\% of the six-task Full
average, while the pathway intervention yields the same positive low-budget
message effect.

LLaVA changes the vision encoder, projector, decoder organization, source token
count, and multimodal positional interface.  GMC uses the corresponding native
Q/K/V projections and source positions without retraining or
architecture-specific parameter adjustment.  At 128 tokens, the six-task
average remains at 99.59\% of Full; at 64 tokens, it remains at 99.68\%.
The latter point preserves Full-level POPE and MMBench behavior while slightly
improving MME and AMBER.

The pathway intervention provides a stronger transfer test than aggregate
retention alone.  With appearance/spatial coverage, assignment, transport
machinery, positions, and $K=64$ fixed, enabling current and historical message
clients plus message-derived population weights raises TextVQA by 4.00 points
and the three-task macro by 1.16 points.  The same low-rate mechanism that
improves Qwen sparse reading therefore remains active under LLaVA's different
visual and language representations.

\section{Additional Ablations and Qualitative Results}
\label{sec:cond-ablations}

\subsection{Components and compact representations}

\begin{table*}[t]
\centering
\small
\begin{minipage}[t]{0.48\linewidth}
\centering
\textit{(a) Marginal component effects at $K=128$}\\[2pt]
\setlength{\tabcolsep}{2.8pt}
\begin{tabular}{@{}lrr@{}}
\toprule
Component & TextVQA $\Delta$ [95\% CI] & ChartQA $\Delta$ [95\% CI] \\
\midrule
Message & $+0.58$ [$-0.20,+1.35$] & $+1.12$ [$-0.56,+2.80$] \\
Appearance & $+8.03$ [$+7.09,+8.97$] & $+8.12$ [$+6.40,+9.84$] \\
Spatial & $-0.95$ [$-1.67,-0.22$] & $-0.60$ [$-2.20,+1.00$] \\
Transport & $+1.28$ [$+0.59,+1.97$] & $+2.64$ [$+1.32,+3.96$] \\
\bottomrule
\end{tabular}

\vspace{5pt}
\textit{(b) Same-support recovery at H2, $K=128$}\\[2pt]
\setlength{\tabcolsep}{1.2pt}
\begin{tabular}{@{}lrrrr@{}}
\toprule
Task & Uniform & Ground. & $\Delta$ [95\% CI] & Err. U/G \\
\midrule
TextVQA & 79.64 & 78.44 & $-1.20$ [$-3.48,+0.92$] & .2916/.2777 \\
ChartQA & 44.00 & 46.00 & $+2.00$ [$-2.40,+6.40$] & .3875/.3404 \\
DocVQA & 53.48 & \textbf{58.39} & \textbf{$+4.91$ [$+1.17,+8.73$]} & .4000/.3269 \\
\bottomrule
\end{tabular}
\end{minipage}\hfill
\begin{minipage}[t]{0.48\linewidth}
\centering
\textit{(c) Spatial bank in the complete matrix}\\[2pt]
\setlength{\tabcolsep}{2.8pt}
\begin{tabular}{@{}llrr@{}}
\toprule
Boundary & Configuration & Macro & $\Delta$ [95\% CI] \\
\midrule
H2 & M+A+T & 76.55 & - \\
H2 & M+A+S+T & 76.47 & $-0.08$ [$-0.63,+0.48$] \\
L16 & M+A+T & 82.09 & - \\
L16 & M+A+S+T & 82.13 & $+0.05$ [$-0.28,+0.37$] \\
\bottomrule
\end{tabular}

\vspace{5pt}
\textit{(d) Original multimodal positions}\\[2pt]
\setlength{\tabcolsep}{3.2pt}
\begin{tabular}{@{}lrrr@{}}
\toprule
ChartQA, $K=256$ & Original & Packed & $\Delta$ [95\% CI] \\
\midrule
Hard support & 70.0 & 64.0 & $-6.0$ [$-12.0,-1.0$] \\
Local refit & 69.0 & 67.0 & $-2.0$ [$-8.0,+4.0$] \\
\bottomrule
\end{tabular}
\end{minipage}
\caption{Component and representation ablations.  Panel (a) adds each named
component to its matched ablation arm.  Panel (b) fixes support indices and
changes only the recovery map; Error is relative boundary-message distortion.
Panel (c) compares the full client matrix with and without the low-mass spatial
bank, and panel (d) replaces source M-RoPE positions by consecutive packed
positions.}
\label{tab:cond-components}
\end{table*}

Appearance coverage supplies the largest isolated gain, while population
transport improves both tasks and lowers boundary-message error in every
same-support recovery row.  Spatial coverage is analyzed in the following subsection; preserving original multimodal
positions remains important when no state recovery can absorb the coordinate
perturbation.

The component effects distinguish bank allocation from representation
realization.  Removing appearance clients causes the largest loss on both
TextVQA and ChartQA, showing that prompt attention alone does not cover all
future evidence.  Message clients provide smaller positive marginal effects in
this local component design because the appearance bank still retains many of
the same answer carriers; the complete-path experiments in
Table~\ref{tab:cond-message-path} expose their larger effect under severe
budgets.  Transport has a resolved positive contribution on both tasks.

The recovery rows hold support fixed at $K=128$.  Grounded transport lowers
relative message error from .4000 to .3269 on DocVQA and produces a
$+4.91$-point ANLS gain over uniform merging.  ChartQA moves in the same
direction, while TextVQA remains statistically tied.  These differences agree
with the evidence geometry: document strings and chart values often span
several neighboring patches whose population should be accumulated by one
representative.

The position intervention changes no support state or recovery rule.  Packed
positions renumber the selected visual sequence consecutively, whereas
original positions retain the source M-RoPE coordinates.  The six-point loss
for hard support shows that feature preservation alone does not preserve the
decoder's spatial reference frame.  Local refit reduces but does not reverse
the loss, so all GMC operating points retain original positions.

\subsection{Spatial Coverage under Distributed Evidence}
\label{sec:cond-spatial-distributed}

The spatial bank is a low-weight prior over original image coordinates.  With
message and appearance clients fixed, its support allocation is the
scalarization
\begin{equation}
 S_{\lambda_s}=\arg\max_{|S|\le K}
 \left\{F_{M+A}(S)+\lambda_sF_S(S)\right\},\;\lambda_s=0.25.
 \label{eq:cond-spatial-scalarization}
\end{equation}
For spatial landmark clients $P_m^S$, the corresponding uncovered-grid
distortion is
\begin{equation}
 D_{\mathrm{grid}}(S)=\frac{1}{M_S}\sum_{m=1}^{M_S}
 \left(1-\max_{i\in S}P_m^S(i)\right).
 \label{eq:cond-grid-distortion}
\end{equation}
Increasing spatial coverage lowers $D_{\mathrm{grid}}$ and reduces support
collapse onto a small prompt-salient region.  At fixed $K$, the same slot
cannot simultaneously expand global coverage and further refine a local text
or chart region, yielding a boundary- and task-dependent trade-off.

Panel (c) of Table~\ref{tab:cond-components} shows that the aggregate effect is
near zero: H2 changes by $-0.08$ points [${-0.63},{+0.48}$], and L16 by
$+0.05$ [${-0.28},{+0.37}$].  The early-boundary effect is concentrated in
MME categories whose evidence is distributed across the image.

\begin{table*}[t]
\centering
\small
\setlength{\tabcolsep}{3.0pt}
\begin{tabular*}{\linewidth}{@{\extracolsep{\fill}}lrrr@{}}
\toprule
Category & Without spatial & With spatial & $\Delta$ [95\% CI] \\
\midrule
MME cognition family & 578.21 & \textbf{599.64} & \textbf{$+21.43$ [$+2.86,+45.36$]} \\
MME perception family & 1646.23 & \textbf{1660.83} & $+14.60$ [$-15.31,+46.66$] \\
MME code reasoning & 112.50 & \textbf{125.00} & $+12.50$ [$+0.00,+30.00$] \\
MME OCR & 185.00 & \textbf{192.50} & $+7.50$ [$+0.00,+22.50$] \\
MME numerical calculation & 132.50 & \textbf{140.00} & $+7.50$ [$+0.00,+22.50$] \\
MME position & 158.33 & \textbf{165.00} & $+6.67$ [$+0.00,+18.33$] \\
\bottomrule
\end{tabular*}
\caption{Early-boundary effect of spatial coverage on distributed-evidence MME
categories.  The paired H2 configurations differ only in the support-side
spatial bank; assignment, transport, original positions, boundary, and
evaluator remain fixed.}
\label{tab:cond-spatial-distributed}
\end{table*}

The spatial bank is nearly neutral on the aggregate, while its early-boundary
effect is concentrated in distributed-evidence categories.  Cognition improves
by 21.43 points, with consistent positive directions on OCR, numerical
calculation, code reasoning, and position.  At L16, deeper contextual mixing
reduces the marginal value of this broad coordinate reserve.

\subsection{Probe and parameter sensitivity}

\begin{table*}[t]
\centering
\small
\begin{minipage}[t]{0.37\linewidth}
\centering
\textit{(a) Grounded-message probe}\\[2pt]
\setlength{\tabcolsep}{3.2pt}
\begin{tabular}{@{}lrr@{}}
\toprule
Projection & Macro & Signed error \\
\midrule
Rank 4, seed 0 & 79.80 & .1869 \\
Rank 32, seed 0 & 79.53 & .1934 \\
Rank 256, seed 0 & 79.72 & .1937 \\
Full basis, rank 3,584 & 79.59 & .1940 \\
Rank 4, seed 1 & 79.87 & .1883 \\
Rank 4, seed 2 & 79.73 & .1859 \\
\bottomrule
\end{tabular}
\end{minipage}\hfill
\begin{minipage}[t]{0.60\linewidth}
\centering
\textit{(b) Parameter neighborhoods}\\[2pt]
\setlength{\tabcolsep}{2.6pt}
\begin{tabular}{@{}llrr@{}}
\toprule
Family / task & Setting & Score & $\Delta$ [95\% CI] \\
\midrule
Population / ChartQA & .25 & 45.00 & $-1.40$ [$-2.80,-.20$] \\
 & .50 (deployed) & 46.40 & $0.00$ [$0.00,0.00$] \\
 & 1.00 & 47.00 & $+.60$ [$-1.60,+2.80$] \\
Spatial assignment / ChartQA & .10 & 46.40 & $-2.80$ [$-5.20,-.40$] \\
 & .25 & 47.40 & $-1.80$ [$-3.80,+.20$] \\
 & .50 (deployed) & 49.20 & $0.00$ [$0.00,0.00$] \\
Ambiguity $(t,p)$ / ChartQA & $(.10,2)$ & 45.80 & $-2.20$ [$-4.00,-.40$] \\
 & $(.10,4)$ & 46.40 & $-1.60$ [$-3.60,+.20$] \\
 & $(.08,8)$ (deployed) & 48.00 & $0.00$ [$0.00,0.00$] \\
Batch / TextVQA & 16 (deployed) & 81.27 & $0.00$ [$0.00,0.00$] \\
 & 64 & 81.80 & $+.53$ [$-2.30,+3.33$] \\
Batch / ChartQA & 16 (deployed) & 64.67 & $0.00$ [$0.00,0.00$] \\
 & 64 & 61.00 & $-3.67$ [$-8.33,+1.00$] \\
\bottomrule
\end{tabular}
\end{minipage}
\caption{Probe and parameter sensitivity on the original paired development
subsets.  Probe rows vary projection rank or seed; parameter rows vary one
family at a time around the deployed setting.  The rank-four probe retains the
task behavior of much wider projections, while the transport and batching
sweeps expose smooth local trade-offs.}
\label{tab:cond-sensitivity}
\end{table*}

The projection sweep varies only the grounded-message probe used during support
construction.  Native unprojected values are used for Signed error in every
row.  Macro varies within 0.34 points from rank four to the full 3,584
dimensional basis, and all three rank-four seeds lie in the same narrow range.
The selected supports need not be identical for downstream behavior to remain
stable: several complementary carrier sets can realize a similar grounded
message when appearance coverage and transport are shared.

The population-strength sweep is locally flat around the deployed .50 setting.
Spatial assignment and ambiguity gating are more visible on ChartQA because
the task couples bars, ticks, legends, and values across positions.  The
responsibility batch controls parallelism: increasing it from 16 to 64 leaves
TextVQA stable but coarsens ChartQA selection on the development subset.  Fixed
resolution therefore uses $b=16$; the long-prefix implementation uses a larger
batch to limit selection rounds after bounding the appearance bank.

\subsection{Qualitative examples}

\begin{figure*}[t]
\centering
\includegraphics[width=\linewidth]{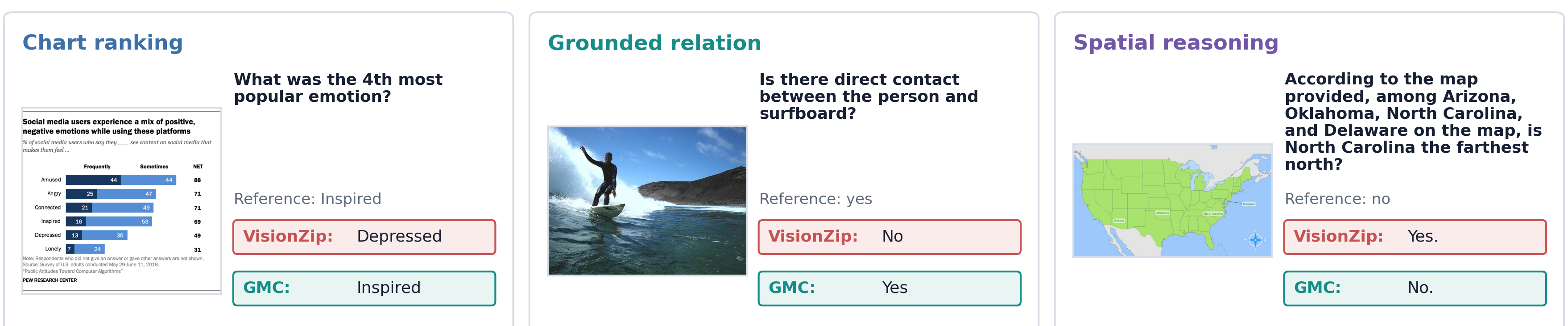}
\caption{Qualitative examples on chart ranking, grounded relation, and map
reasoning.  GMC preserves complementary evidence required by each question
and produces the reference answer under physical token compaction.}
\label{fig:cond-qualitative}
\end{figure*}

Figure~\ref{fig:cond-qualitative} illustrates three complementary evidence
patterns.  Chart ranking requires the label, ordering, and associated value;
grounded relation requires both entities and their interaction; map reasoning
combines text with spatial placement.  GMC preserves the joint evidence needed
for each decision under the compact visual budget.

\section{Mechanism Diagnostics}

Mechanism measurements follow the two interfaces in the method.  Facility
coverage evaluates support allocation before transport.  Signed boundary
messages evaluate the compact representation consumed by the decoder.  Visual
innovation and candidate margins evaluate the downstream effect on image-induced
answer ordering.

For answer position $t$, let $z_t^F$, $z_t^S$, and $z_t^0$ be Full, compact,
and visual-null next-token logits under a shared prompt history.  Define
\begin{equation}
 C_t^{\rm AP}=\operatorname{Top}_{32}(z_t^F)
 \cup\operatorname{Top}_{32}(z_t^0),\qquad
 H_C=I-|C|^{-1}\mathbf1\mathbf1^\top .
 \label{eq:cond-vie-candidates}
\end{equation}
For $E\in\{F,S\}$, the centered visual innovation is
$\nu_t^E=H_C(z_{t,C}^E-z_{t,C}^0)$.  Let
$\bar z^E=H_Cz_{t,C}^E$ and freeze diagonal anti-prior weights
\begin{equation}
 [W_{F0}]_{jj}=1+
 \operatorname{sigm}(-\bar z_j^F\bar z_j^0)
 |\bar z_j^F-\bar z_j^0|.
 \label{eq:cond-vie-weights}
\end{equation}
The operational visual-innovation error is
\begin{equation}
 {\rm VIE}_t=
 \frac{\norm{W_{F0}^{1/2}(\nu_t^F-\nu_t^S)}_2}
 {\max\{\norm{W_{F0}^{1/2}\nu_t^F}_2,10^{-8}\}}.
 \label{eq:cond-vie}
\end{equation}
The Full/null pair fixes the candidate set and weights before compact logits
are inspected.  VIE is evaluated after compression and is not used to select
tokens.

For signed boundary fidelity, stack query/head messages with their evaluation
weights into $\mathsf Y$ and $\widetilde{\mathsf Y}$.  Direction
$\kappa=\langle\mathsf Y,\widetilde{\mathsf Y}\rangle/
(\norm{\mathsf Y}_2\norm{\widetilde{\mathsf Y}}_2)$ and amplitude
$\alpha=\norm{\widetilde{\mathsf Y}}_2/\norm{\mathsf Y}_2$ obey
\begin{equation}
 \frac{\norm{\mathsf Y-\widetilde{\mathsf Y}}_2^2}
 {\norm{\mathsf Y}_2^2}=1+\alpha^2-2\alpha\kappa.
 \label{eq:cond-direction-magnitude}
\end{equation}
This identity records cancellation and scale that cannot be inferred from
nonnegative facility energy.  Grouped attention-mass error is reported
separately because compact keys and positions change softmax normalization.

The support and output diagnostics use fixed samplewise definitions.  For bank
$b$, let $F_b$ denote its facility contribution and define saturation
\begin{equation}
 c_b(S)=\frac{F_b(S)}{F_b(V)},\qquad
 c(S)=\frac{F(S)}{F(V)}=\sum_b\omega_bc_b(S),
 \label{eq:cond-saturation}
\end{equation}
with $\omega_b=F_b(V)/F(V)$.  This ratio is computed before transport and
measures how much of the available client geometry the returned support covers.
The signed-message diagnostic is computed after support and recovery by
re-running native compact attention in the unprojected value space.  Its
relative error is
\begin{equation}
 e_M=\frac{\norm{\mathsf Y-\widetilde{\mathsf Y}}_2}
 {\norm{\mathsf Y}_2+10^{-8}},
 \label{eq:cond-message-error}
\end{equation}
with query/head weights identical to those used by the corresponding boundary
measurement.

For a task example $x$, let $c_x^F$ indicate that Full is correct and let
$a_x^F,a_x^S$ be the normalized Full and compact answers.  The harmful-change
indicator is
\begin{equation}
 h_x=\mathbf1[c_x^F=1]\mathbf1[a_x^S\ne a_x^F].
 \label{eq:cond-harmful}
\end{equation}
AUROC evaluates whether VIE orders these events above unchanged Full-correct
examples.  Correlations use continuous, centered candidate-logit distortion
rather than the binary event and are grouped by question when several
operating points share an input.  These quantities test different stages:
$c_b$ measures support, $e_M$ measures realized attention messages, VIE
measures image-induced output displacement, and $h_x$ records a task-level
failure.

The output certificate normalizes distortion by the Full decision margin.
For
$C_x^{\rm win}=\operatorname{Top}_{32}(z_x^F)\cup
\operatorname{Top}_{32}(z_x^S)$, define
\begin{equation}
 D_x=\norm{H_{C_x^{\rm win}}(z_x^F-z_x^S)}_2,qquad
 \rho_x=\frac{\sqrt2D_x}{\Gamma_x^F},
 \label{eq:cond-certificate-ratio}
\end{equation}
where $\Gamma_x^F$ is Full's vocabulary-wide top-two margin.  Because this set
contains both observed winners, $\rho_x<1$ is a deterministic sufficient
condition for winner agreement.  Its empirical coverage is reported
separately from the prospective Full/null candidate-set check in
Eq.~\eqref{eq:cond-vie-candidates}.

\begin{table*}[t]
\centering
\small
\begin{minipage}[t]{0.51\linewidth}
\centering
\textit{(a) Final-support coverage: mean / fifth percentile}\\[2pt]
\setlength{\tabcolsep}{0.4pt}
\begin{tabular}{@{}lrrrrr@{}}
\toprule
Boundary & $K/N$ & All & Ground. & Appear. & Spatial \\
\midrule
Qwen H2 & 128/1,296 & .942/.895 & 1.000/.999 & .437/.389 & .914/.906 \\
Qwen L16 & 128/1,296 & .923/.888 & .995/.985 & .387/.342 & .899/.880 \\
LLaVA L16 & 64/576 & .929/.903 & .994/.988 & .456/.387 & .827/.804 \\
Doc. H1 & 4096/15876 & 1.000/1.000 & 1.000/1.000 & .967/.959 & .999/.999 \\
\bottomrule
\end{tabular}

\vspace{5pt}
\textit{(b) Visual-innovation diagnostics}\\[2pt]
\setlength{\tabcolsep}{3.0pt}
\begin{tabular}{@{}lrrrr@{}}
\toprule
Method & VIE$\downarrow$ & Harmful$\downarrow$ & AUC$\uparrow$ & $r$(logit,VIE) \\
\midrule
Norm+Sim & .456 & 35.7\% & .770 & - \\
Head-aware control & .312 & 21.7\% & .832 & - \\
GMC hard & .195 & 10.9\% & \textbf{.855} & - \\
GMC transport & \textbf{.192} & \textbf{9.2\%} & .826 & .887 \\
\bottomrule
\end{tabular}
\end{minipage}\hfill
\begin{minipage}[t]{0.45\linewidth}
\centering
\textit{(c) Output-interface summaries}\\[2pt]
\setlength{\tabcolsep}{0.8pt}
\begin{tabular}{@{}lrrr@{}}
\toprule
Quantity & Estimate & 95\% CI / events & Viol. \\
\midrule
No-flip certificate & 21.6\% & [18.8,24.5] & 0 \\
Candidate check & 450 & 27 flips & 0 escapes \\
\bottomrule
\end{tabular}

\vspace{7pt}
The coverage ratio
\[
 \widehat c(S)=F(S)/F(V)
 \leq F(S)/F(S^\star)
\]
is a per-example lower bound on achieved facility coverage relative to an
optimal size-$K$ support.  It concerns the support objective; the signed-message
and output diagnostics test the subsequent realization and decision interfaces.

The candidate set for the escape study is fixed before compact logits are
inspected.  Across 450 operating-point conditions, every compact winner remains
inside that set, including all 27 observed winner changes.
\end{minipage}
\caption{Mechanism diagnostics.  Coverage uses the selected support before
transport.  VIE is centered visual-innovation error; Harmful is the rate of
Full-correct answers changed by compression; AUC measures VIE's discrimination
of those changes.  The deterministic winner-margin condition produces a
non-vacuous no-flip certificate on 21.6\% of evaluated cases with no observed
violations.}
\label{tab:cond-mechanism}
\end{table*}

Grounded clients are nearly saturated even at the narrowest budgets, while
appearance coverage remains the active rate-distortion term.  GMC support
reduces both VIE and harmful answer changes relative to independent feature and
attention controls.  The stronger correlation at candidate logits is
consistent with the propagation of an upstream message perturbation toward the
decision boundary.

Panel (a) summarizes four qualitatively different operating points.  Grounded
coverage is at least .985 at the fifth percentile for all four, whereas
appearance coverage is lower at narrow fixed-token budgets and nearly saturated
for the 4,096-token document support.  This identifies appearance geometry as
the principal finite-rate constraint after query-addressed message modes have
been retained.  Spatial saturation remains high despite its low objective mass.

Panel (b) evaluates methods after their compact execution.  Moving from
independent norm/similarity selection to head-aware attention reduces VIE and
harmful changes; GMC hard support reduces both further.  Population transport
achieves the lowest mean VIE and harmful-flip rate.  Candidate-logit distortion
correlates strongly with VIE, linking the image-induced ranking diagnostic to
the final decoder boundary.

The no-flip condition uses Full's candidate margin and the centered Full-GMC
logit distortion.  A condition is counted as certified when the deterministic
bound in Eq.~\eqref{eq:cond-margin-bound} is below the represented Full margin.
The condition applies to 21.6\% of evaluated cases and has zero observed
violations.  Separately, the prospective candidate set in
Eq.~\eqref{eq:cond-vie-candidates} contains every compact winner in 450 tested
conditions, including all 27 observed top-1 changes.  These measurements cover
candidate inclusion and ordering; the empirical fraction is reported
separately from the analytical Lipschitz decomposition.

\section{Additional Method Details}
\label{sec:cond-method}

This section specifies native message construction, complementary support,
population transport, physical compaction, and fixed-support decoding.  The
shared deployed constants appear in Table~\ref{tab:cond-config}.

\subsection{Native message clients}

Let $H^{p-1}\in\R^{L\times d}$ be the residual stream immediately before
decoder block $p$, with text positions $\mathcal T$ and visual positions
$\mathcal V$.  Using the block's native RMS normalization, multimodal rotary
positions, causal mask, and projections gives
\begin{align}
 Q_h&=X_{\mathcal T}W_h^Q,&
 K_h^{\rm all}&=X_{\le\mathcal T}W_h^K,\notag\\
 V_h&=X_{\mathcal V}W_h^V,\notag\\
 S_h&=(Q_hK_h^{{\rm all}\top}+M_{\rm causal})/\sqrt{d_h},\notag\\
 A_h^{\rm all}&=\operatorname{softmax}(S_h),\notag\\
 A_h&=A_h^{\rm all}[:,\mathcal V],&
 Y_h&=A_hV_h.
 \label{eq:cond-native-message}
\end{align}
Support construction therefore approximates the visual-column message that
the next block will consume.  It does not renormalize a separate visual-only
attention matrix.

The boundary is an execution interface, not a post-hoc saliency probe.  Blocks
$0{:}p-1$ run on the complete visual sequence, and the clients are built with
the normalization and projections of the next block to execute.  The selected
and transported states then replace the visual columns before block $p$.
Consequently, the client attention includes the native nonvisual denominator
and the original causal geometry.  Computing a softmax only over visual columns
would allocate unit mass to the image regardless of how strongly the prompt
actually attends to it; Eq.~\eqref{eq:cond-native-message} retains the visual
mass assigned by the model.

For H1, H2, and L16, the current view is evaluated at the corresponding
compression boundary and remains distinct from completed historical views.
With a history count of two, the completed view sets are $\{0\}$, $\{0,1\}$,
and $\{14,15\}$, respectively.  A historical view preserves evidence already
integrated by an earlier block even when the current layer redistributes its
attention.  Each view remains a separate collection of clients, so history
adds coverage constraints rather than averaging potentially specialized
messages into one score.

For current and two completed historical views $c$, query $t$, head $h$, and
visual token $i$, let $Z_{c,h}=(V_{c,h}O_h)P\in\R^{N\times r}$, where $P$ is
a deterministic rank-four Gaussian-QR probe shared by all heads and samples.
The homogeneous attention-mass coordinate is
\begin{align}
 \sigma_{c,h}&=\sqrt{\lambda_{\rm mass}}
 \max\!\left\{\norm{Z_{c,h}}_F/\sqrt{Nr},
 \varepsilon_{\rm mass}\right\},\notag\\
 \lambda_{\rm mass}&=1,\qquad\varepsilon_{\rm mass}=10^{-8}.
 \label{eq:cond-mass-coordinate}
\end{align}
The projected grounded atom and normalized client distribution are
\begin{align}
 u_i^{c,h,t}&=A_{c,h}[t,i]
 [(V_{c,h}[i]O_h)P;\sigma_{c,h}],\notag\\
 P_{c,h,t}^{G}(i)&=
 \frac{\norm{u_i^{c,h,t}}_2}{\sum_j\norm{u_j^{c,h,t}}_2}.
 \label{eq:cond-grounded-client}
\end{align}
The appended coordinate retains visual attention mass on the same headwise
scale as the projected value.  Current and historical query rows remain
separate.  The terminal prompt query receives twice the weight of other query
positions in message energy and transport statistics.  Head contribution
patterns are clustered into at most four groups; clusters receive equal total
mass, and inverse-dispersion weights preserve specialized heads within each
cluster.

The probe is constructed once on CPU: seed zero draws
$G\in\R^{d\times4}$ with independent standard-normal entries and reduced QR
factorization $G=PR$ supplies the orthonormal columns of $P$.  Each head value
is first mapped by its native output-projection block $O_h$ and then by $P$.
The rank-four projection is used only to allocate support; all signed-message
diagnostics recompute native, unprojected head values.  This distinction is
important because support construction needs a compact representation of many
head-query contributions, whereas the downstream diagnostic should measure
the actual vector message consumed by the model.

To avoid head-count bias, we cluster normalized per-head contribution patterns
with farthest-first cosine clustering.  Let $\mathcal H_c$ denote a cluster,
$\mu_c$ its normalized mean pattern, and
$d_h=1-\langle \widehat b_h,\mu_c\rangle$ the within-cluster dispersion.
Cluster $c$ receives equal total mass; its members receive weights proportional
to $(d_h+\epsilon)^{-1}$ and are renormalized within the cluster.  A small OCR-
or relation-specialized cluster can therefore contribute comparably to a large
family of redundant heads.  No task label or parameter update enters this
construction.

More precisely, let $U_h^v$ collect the flattened atoms of head $h$ in view
$v$, normalized to unit average atom norm, and let $\beta_v$ be that view's
mass.  The pattern used for clustering is
\begin{equation}
 u_h=\frac{\operatorname{concat}_v
 [\sqrt{\beta_v}\operatorname{vec}(U_h^v)]}
 {\norm{\operatorname{concat}_v
 [\sqrt{\beta_v}\operatorname{vec}(U_h^v)]}_2}.
 \label{eq:cond-head-pattern}
\end{equation}
Farthest-first initialization chooses the largest pre-normalization pattern,
then repeatedly adds the head with smallest maximum cosine to existing centers.
After assigning each head to its nearest center, the deployed weight is
\begin{equation}
 \omega_h=\frac1{C}\,
 \frac{(d_h+0.05)^{-1}}
 {\sum_{g\in\mathcal H_{c(h)}}(d_g+0.05)^{-1}},
 \qquad d_h=1-u_h^\top\mu_{c(h)},
 \label{eq:cond-head-weight}
\end{equation}
where $C$ is the number of nonempty clusters.  Every cluster receives mass
$1/C$, while internally consistent members receive a larger share within that
cluster.  An exactly zero-energy query/head row remains inactive instead of
being converted into a uniform visual distribution.

\subsection{Appearance and spatial clients}

Grounded clients cover evidence already accessed by the prompt.  Appearance
and spatial clients retain complementary visual and coordinate structure:
\begin{align}
 P^V_m(i)&\propto\exp(x_m^\top x_i/\tau_v),\notag\\
 P^S_m(i)&\propto\mathbf1[I_i=I_m]
 \exp[-\norm{p_i-\ell_m}_2^2/\tau_s].
 \label{eq:cond-aux-clients}
\end{align}
Here $x_i$ is the normalized boundary state, $p_i$ is the original image-grid
coordinate, and $I_i$ is the image index.  Positive-energy rows are
normalized before their bank mass is applied.  For bank $b$, potential client
$m$, active-row indicator $a_m$, and nominal mass $\beta_b$,
\begin{equation}
 \bar C_{mi}=\frac{\beta_b}{M_b}a_mP_m(i),\qquad
 F(S)=\sum_m\max_{i\in S}\bar C_{mi}.
 \label{eq:cond-coverage}
\end{equation}
$F$ is normalized, monotone, and submodular.  It values a token only through
the residual client mass it adds to the selected set, rather than through an
independent scalar ranking.

The nominal bank mass and its effective facility share are distinct.  For bank
$b$ with active-row indicator $a_m$, define
\begin{align}
 \widehat\beta_b&=\beta_bM_b^{-1}\sum_{m\in b}a_m,\notag\\
 Z_b&=F_b(V)=\frac{\beta_b}{M_b}
 \sum_{m\in b}a_m\max_iP_m(i),\notag\\
 \omega_b&=Z_b/\sum_cZ_c.
 \label{eq:cond-effective-bank-mass}
\end{align}
$\widehat\beta_b$ records deployed matrix mass, while $\omega_b$ includes the
peakiness of active clients and is the coefficient in the exact saturation
mixture of Eq.~\eqref{eq:cond-saturation}.  Appearance and spatial rows are
strictly positive; grounded rows can be inactive when the native visual
contribution is zero.

The three banks encode different coverage geometries.  A grounded row asks
whether the support contains a carrier for a message already addressed by the
prompt.  An appearance row asks whether each boundary state has a nearby
representative in cosine geometry, including visually meaningful states that
have not yet received strong prompt attention.  A spatial row places a small
amount of mass on coarse original-grid landmarks.  It prevents a support from
collapsing entirely onto one salient region and preserves anchors for spatial
relations.  Their masses are applied after row normalization, so the number of
clients in one bank cannot increase its total weight merely by replication.

For multiple images, appearance and spatial rows are restricted to tokens from
their own image.  The $16\times16$ landmarks are defined in normalized source
coordinates and mapped to each image grid; selected tokens retain the complete
native multimodal position tuple.  Spatial clients therefore influence which
source positions remain available but never introduce synthetic positions.

The fixed $N=1{,}296$ path uses every normalized boundary state as an
appearance client.  For long prefixes, appearance states are projected by a
fixed rank-128 Rademacher matrix and at most $M_V=4{,}096$ uniformly spaced
clients are retained.  This bounds appearance memory by $O(M_VN)$.

\subsection{Batched complementary support}

For requested budget $K_r$ and $I$ input images, the deployed budget is
$K=\min\{N,\max(K_r,I)\}$.  Multi-image prompts are initialized with one
maximum-gain token per image.  At support $S_t$, define residual responsibility
\begin{align}
 R_{mi}^t&=[\bar C_{mi}-\max_{j\in S_t}\bar C_{mj}]_+,\notag\\
 \rho_{mi}^t&=\operatorname{softmax}_{i\in\mathcal P_t}
 \left(\frac{R_{mi}^t}{0.1\max_{j\in\mathcal P_t}R_{mj}^t+\epsilon}\right),\notag\\
 s_i^t&=\sum_m\rho_{mi}^tR_{mi}^t.
 \label{eq:cond-responsibility}
\end{align}
$\mathcal P_t$ contains the $4b$ largest singleton gains, and the top $b=16$
responsibility scores enter together.  Scores are recomputed after each batch,
so clients already covered by one representative cease to attract redundant
tokens.

The candidate pool limits the expensive responsibility normalization without
turning the solver into a fixed top-$K$ ranking.  Singleton gain first excludes
tokens that cannot improve the current support.  Within the remaining pool,
$\rho_{mi}^t$ distributes each uncovered client's residual among its plausible
representatives; summing $\rho_{mi}^tR_{mi}^t$ favors tokens that assume
responsibility for different client groups.  After a batch enters, all maxima
and residuals are refreshed.  The support trajectory is thus conditional on
what has already been represented.

The totality rule handles edge cases deterministically.  A multi-image prompt
is seeded with one maximum-gain token per image.  If $K_r<I$, the effective
budget expands to $I$; if $K_r>N$, it clamps to $N$; and $K=N$ returns the
unmodified execution.  When all remaining gains are zero, stable source-index
order fills the exact physical budget without changing $F$.  These rules keep
assignment image-compatible and make every requested setting well defined.

\subsection{Population assignment and transport}

Each source token is assigned to an image-compatible representative using
boundary-state cosine similarity and an ambiguity-gated spatial term.  With
$\hat h_i=h_i/\max\{\norm{h_i}_2,\epsilon\}$, original-grid coordinate $p_i$,
sample gate $s_x$, and local ambiguity gate $g_i$,
\begin{align}
 \kappa_{ij}&=\hat h_i^\top\hat h_j+0.5s_xg_i
 \exp[-\norm{p_i-p_j}_2^2/\tau_s],\notag\\
 \pi(i)&=\arg\max_{j\in S_i}\kappa_{ij}.
 \label{eq:cond-assignment}
\end{align}
Assignments cannot cross images, and selected tokens represent themselves.
For cluster $\mathcal C_j=\{i:\pi(i)=j\}$ and positive message-derived mass
$m_i$,
\begin{align}
 M_j&=\sum_{i\in\mathcal C_j}m_i,
 &c_j&=M_j^{-1}\sum_{i\in\mathcal C_j}m_ih_i,\notag\\
 \widetilde h_j&=\mathcal R_j(c_j),
 &\mathcal R_j(u)&=u\,
 \frac{\operatorname{rms}(h_j)}{\operatorname{rms}(u)+\epsilon}.
 \label{eq:cond-transport}
\end{align}
Grounded energy sets $m_i$ through the frozen population-strength gate given
in Table~\ref{tab:cond-config}.  Before RMS restoration, transport preserves
the exact signed first moment for every linear map $W$:
\begin{equation}
 \sum_{j\in S}M_jWc_j
 =\sum_{j\in S}W\!\left(\sum_{i\in\mathcal C_j}m_ih_i\right)
 =\sum_{i=1}^{N}m_iWh_i.
 \label{eq:cond-population-moment}
\end{equation}
The compact sequence keeps each representative's original multimodal position
ID and leaves all text positions unchanged.

The message-derived population weights are computed from native current-view
energy.  Let
\begin{equation}
 e_i=\left(\sum_h\omega_h
 \norm{r_i^{{\rm current},h,\cdot}}_2^2\right)^{1/2},
 \qquad
 \bar e_i=\frac{e_i}{N^{-1}\sum_j e_j}.
 \label{eq:cond-grounded-energy}
\end{equation}
Let $\bar\Delta_x$ be the mean gap between the largest and second-largest
compatible selected-state cosine similarities over discarded tokens.  The
sample-separability gate and population weight are
\begin{align}
 s_x&=1-\exp[-(\bar\Delta_x/0.08)^8],\notag\\
 m_i&=(1-0.5s_x)+0.5s_x\bar e_i.
 \label{eq:cond-population-gate}
\end{align}
When a compatible set has one representative its gap is zero; when no token is
discarded, $\bar\Delta_x=0$; and if the energy normalizer vanishes,
$\bar e_i=1$.  These conventions make the endpoint total.  Ambiguous examples
remain close to uniform population transport, while separable examples use
message energy more strongly.  The same $s_x$ scales the spatial term in
Eq.~\eqref{eq:cond-assignment}.

RMS restoration preserves the selected carrier's native residual-stream scale
after its direction has been replaced by the population centroid.  It does not
change the representative's source position.  Thus each compact token consists
of three coordinated pieces: transported state, selected carrier scale, and
original M-RoPE position.  The representation ablations above vary recovery and positions separately.

\begin{table*}[t]
\centering
\small
\setlength{\tabcolsep}{2.4pt}
\begin{tabular*}{\linewidth}{@{\extracolsep{\fill}}p{0.18\linewidth}p{0.32\linewidth}p{0.43\linewidth}@{}}
\toprule
Component & Setting & Scope \\
\midrule
Message construction & rank 4, Gaussian-QR seed 0; $\lambda_{\rm mass}=1$, $\varepsilon_{\rm mass}=10^{-8}$ & All fixed-token runs \\
Message views & current/history mass $0.5/0.5$; two historical views & Qwen and LLaVA quality experiments \\
Auxiliary banks & appearance mass $0.5$, spatial mass $0.25$; $\tau_v=0.2$, $\tau_s=0.02$ & Shared across tasks and budgets \\
Spatial clients & $16\times16$ original-grid landmarks & Low-weight coordinate coverage prior \\
Solver & $b=16$, candidate pool $4b$, temperature $0.1$ & Fixed $N=1{,}296$ evaluation \\
Assignment & spatial weight $0.5$; gate $(0.08,8)$; margin $0.05$ & Same-image population assignment \\
Transport & grounded population strength $0.5$; RMS restoration & All GMC transport rows \\
Sequence interface & original multimodal position IDs; stable index ties & Every architecture and boundary \\
Long prefix & appearance rank 128, seed 104729, $M_V\le4{,}096$, $b=512$ & $N=15{,}876$ document path \\
\bottomrule
\end{tabular*}
\caption{Shared GMC configuration.  The fixed-token settings transfer to
LLaVA-1.5 without architecture-specific retuning; the long-prefix row changes
only bounded-capacity handling.}
\label{tab:cond-config}
\end{table*}

\subsection{Algorithm and complexity}

\begin{figure*}[t]
\centering
\begin{minipage}{0.98\linewidth}
\small
\hrule
\noindent\textbf{Algorithm 1: Grounded Message Coreset inference}\par
\noindent\textbf{Input:} image-prompt sequence, decoder boundary $p$,
visual positions $\mathcal V$, original multimodal positions $P$, and budget
$K_r$.  Set $K\leftarrow\min\{N,\max(K_r,I)\}$.\par
\begin{minipage}[t]{0.485\linewidth}
\textbf{1. Dense boundary pass.} Encode the image and run blocks
$0{:}p-1$ with all $N$ visual tokens.  If $K=N$, return this execution.\par
\textbf{2. Native clients.} Reuse block $p$'s normalized Q/K/V projections
to construct grounded message, appearance, and spatial clients.\par
\textbf{3. Complementary support.} Initialize one support token per image,
then apply Eq.~\eqref{eq:cond-responsibility} until $|S|=K$; zero-gain
remainders use stable index order.\par
\textbf{4. Population assignment.} Assign all source tokens with
Eq.~\eqref{eq:cond-assignment}, enforcing same-image clusters and self
assignment for selected tokens.\par
\end{minipage}\hfill
\begin{minipage}[t]{0.485\linewidth}
\textbf{5. Transport.} Compute message-derived masses, weighted centroids,
and RMS-restored representatives using Eq.~\eqref{eq:cond-transport}.\par
\textbf{6. Physical compaction.} Replace selected boundary states by
$\widetilde H_S$, delete unselected visual positions, and retain original
multimodal position IDs.\par
\textbf{7. Remaining inference.} Run blocks $p{:}L$ on the compact sequence.
Support, assignments, and transported states are computed once per prompt.\par
\textbf{Output:} generated answer and a physical $K$-token visual coreset.\par
\end{minipage}
\hrule
\end{minipage}
\caption{End-to-end GMC execution.  Selection occurs before block $p$ and
reuses its native projections; no auxiliary model or second complete decode is
introduced.}
\label{alg:cond-gmc}
\end{figure*}

For $M$ clients, $N$ source tokens, responsibility batch $b$, and support $K$,
selection uses $\lceil K/b\rceil$ tensorized rounds.  The fixed-resolution
appearance bank is an exact $N\times N$ block; the bounded long-prefix path
uses $M_G+M_V+M_S$ clients and chunked $M\times N$ gains.  Assignment uses a
chunked $N\times K$ product, and transport uses scatter reductions.  With
$L$ decoder blocks, pruning before block $p$ retains
\begin{equation}
 W(p,K)=pN+(L-p)K,\qquad K_{\rm eff}=W(p,K)/L
 \label{eq:cond-work}
\end{equation}
visual token-layer work.  Final $K$ controls compact prompt and prompt-KV
capacity; $W(p,K)$ accounts for the dense prefix as well.

The fixed-resolution path materializes an exact $N\times N$ appearance block.
With $M$ total clients, one residual update costs $O(MN)$ and the deployed
solver performs $\lceil(K-|S_0|)/b\rceil$ tensorized rounds.  Candidate pooling
reduces the responsibility softmax from all $N$ columns to at most $4b$ without
changing the singleton-gain computation.  Assignment is a chunked $N\times K$
cosine product and transport consists of indexed weighted sums.  The dominant
selection storage is the client matrix at fixed resolution; hidden-state
transport itself is $O(Nd)$ storage and arithmetic.

The long-prefix path keeps the same objective while bounding its appearance
capacity.  A fixed rank-128 Rademacher projection maps the boundary states,
at most $M_V=4{,}096$ uniformly spaced source states become appearance clients,
and client gains are evaluated in chunks.  Its working memory is
$O((M_G+M_V+M_S)N)$ rather than $O(N^2)$, while assignment remains chunked.
The selected support, transported states, and exact source positions are
identical interfaces in the two paths.

\paragraph{Boundary-attention implementation.}
GMC does not materialize the decoder's complete sequence-by-sequence attention
matrix.  At the pruning boundary it projects only question-text queries against
visual keys and retains an $H\times Q\times N$ visual-column slice, where $Q$
is the number of prompt query positions used by the clients.  The resulting
time is $O(HQNd_h)$ and transient attention storage is $O(HQN)$.  The native
nonvisual softmax denominator is recovered from the same score computation, so
the slice equals the corresponding columns of full self-attention rather than a
visual-only renormalization.

The exact $N\times N$ object at fixed resolution is the appearance-client bank,
not decoder attention.  At $N=1{,}296$ it is materialized because it is small
and reused across selection rounds.  Long-prefix execution caps the client
count at $C\le4{,}096$, projects appearance states to 128 dimensions, and
evaluates gains in token chunks, reducing this term to $O(CN)$.  Spatial
clients require $O(M_SN)$ with fixed $M_S$, and both population assignment and
state aggregation stream over source-token chunks.

The implementation therefore requires access to boundary Q/K/V and the
question-to-visual attention slice.  Controlled latency rows use the same SDPA
backend for every method so this access cost is included in GMC's measured
message-construction component.  Q/K/V tensors produced for support
construction are reused at the split boundary where possible; no second model
forward or gradient computation is invoked.

\paragraph{Index map and physical token accounting.}
Let $\iota:[1,L]\rightarrow[1,L-K+N]$ map original sequence positions to the
compacted sequence after deleting $\mathcal V\setminus S$.  The gather is
stable, so $\iota$ is order preserving over all retained positions.  Text
positions keep their hidden states and logical position IDs; a retained visual
position receives $\widetilde h_j$ but keeps the complete original M-RoPE
coordinate of $j$.  Prefix-layer caches are gathered by the same original
sequence indices.  The physical visual count is therefore exactly $K$ in every
post-boundary block, and Eq.~\eqref{eq:cond-work} follows by summing $N$ visual
states over the first $p$ blocks and $K$ over the remaining $L-p$ blocks.

In no-cache quality evaluation, each generated step replays dense blocks
$0{:}p-1$ on that method's own history, inserts the fixed prompt-derived
compression intervention, and runs the remaining blocks.  The cached
deployment gathers the prompt K/V entries at selected sequence indices for
prefix blocks and uses transported states from boundary $p$ onward.  The two
executions therefore share the prompt-terminal compact representation while
using their natural cache semantics during autoregressive generation.
Cached and no-cache executions agree on approximately 98\% of normalized answers.  Selection occurs once per prompt in both paths.

\subsection{Fixed-support autoregressive execution}

The coreset is a prompt-time object.  Let
$\mathcal G(I,q)=(S,\widetilde H_S)$ denote deterministic support construction
and transport for image $I$ and prompt $q$.  Let $\Phi_{<p}(I,h)$ run the
vision stack and decoder blocks $0{:}p-1$ on text history $h$, let
$\mathcal C_{S,\widetilde H_S}$ replace the selected visual boundary states,
delete the complement, and retain the selected source positions, and let
$\Psi_{p:L}$ map the compact boundary to next-token logits.  The no-cache
quality path is the own-history recurrence
\begin{align}
 (S,\widetilde H_S)&=\mathcal G(I,q),\notag\\
 B_t^S&=\mathcal C_{S,\widetilde H_S}
 \!\left(\Phi_{<p}(I,[q;y^S_{<t}])\right),\notag\\
 z_t^S&=\Psi_{p:L}(B_t^S,[q;y^S_{<t}]),\qquad
 y_t^S=\operatorname*{arg\,max}_v z_{t,v}^S .
 \label{eq:cond-no-cache-rollout}
\end{align}
The generated prefix can change at every step, but $S$, the assignment, and
$\widetilde H_S$ remain fixed.  Full follows the same recurrence without
$\mathcal C_{S,\widetilde H_S}$.  While the two generated histories agree,
their dense prefix states before the compression boundary are identical in the
no-cache path; after the first output divergence, each execution continues on
its own history.

The cached path constructs the same prompt-time support and compact boundary,
then gathers dense prompt K/V from blocks $0{:}p-1$ at the retained sequence
indices and caches the compact suffix produced by blocks $p{:}L$.  If
$\mathcal K_t^S$ denotes this cache, later decisions obey
\begin{equation}
 \begin{aligned}
 (z_t^S,\mathcal K_{t+1}^S)
 &=\operatorname{Decode}_S(y_{t-1}^S,\mathcal K_t^S),\\
 y_t^S&=\operatorname*{arg\,max}_v z_{t,v}^S .
 \end{aligned}
 \label{eq:cond-cached-rollout}
\end{equation}
Prompt K/V is not recomputed after prefill.  Prefix layers contain gathered
dense selected entries because transport is introduced at boundary $p$;
boundary and suffix layers consume transported representatives.  This division
preserves the dense computation that has already occurred and applies the
coreset exactly where the remaining compact computation begins.

Both paths retain the source multimodal positions of $S$.  Sequence compaction
uses a stable gather over all nonvisual positions and selected visual indices,
so text order, image identity, and the complete M-RoPE coordinate tuple remain
unchanged.  The no-cache path provides a direct functional specification of
Eq.~\eqref{eq:cond-no-cache-rollout}; the cached path implements the same fixed
prompt intervention with ordinary autoregressive cache reuse.  Their measured agreement quantifies the numerical effect of these cache semantics.

\section{Proofs}

\subsection{Submodularity and batched progress}

For a client $m$, let $f_m(S)=\max_{i\in S}\bar C_{mi}$ and
$f_m(\varnothing)=0$.  If $A\subseteq B$ and $i\notin B$, then
\begin{align}
 f_m(A\cup\{i\})-f_m(A)
 &=[\bar C_{mi}-f_m(A)]_+,\notag\\
 &\ge [\bar C_{mi}-f_m(B)]_+.
 \label{eq:cond-submodular}
\end{align}
Thus every $f_m$, and their nonnegative sum $F$, is normalized, monotone, and
submodular.

For reference, sequential greedy is recovered as the special case $b_t=1$ and
chooses the largest singleton marginal.  If $S_t$ contains $t$ elements, the
optimal residual obeys
\begin{equation}
 F(S^\star)-F(S_t)
 \le\sum_{i\in S^\star\setminus S_t}
 [F(S_t\cup\{i\})-F(S_t)].
 \label{eq:cond-greedy-residual}
\end{equation}
At least one remaining singleton therefore gains
$[F(S^\star)-F(S_t)]/K$.  The usual recurrence gives
$F(S_K)\ge[1-(1-1/K)^K]F(S^\star)\ge(1-e^{-1})F(S^\star)$.
This guarantee describes exact greedy; the following argument uses quantities
measured from the deployed batches.

The deployed solver selects batches.  Let $S_t$ be the support before round
$t$, $B_t$ a batch of size $b_t$, and $\delta^t_{(j)}$ the $j$th-largest
remaining singleton marginal.  Submodularity yields
\begin{equation}
 F(S^\star)-F(S_t)\le U_t:=\sum_{j=1}^{K}\delta^t_{(j)}.
\end{equation}
For $U_t>0$, define
\begin{equation}
 \alpha_t=\frac{K}{b_t}\frac{F(S_t\cup B_t)-F(S_t)}{U_t},
 \qquad q_t=\alpha_tb_t/K.
\end{equation}
The realized batch removes at least the fraction $q_t$ of the current optimal
residual.  Iteration gives
\begin{equation}
 F(S_T)\ge
 \left[1-\prod_{t=0}^{T-1}(1-q_t)\right]F(S^\star).
 \label{eq:cond-batched-bound}
\end{equation}
If $U_t=0$, all remaining marginals are zero and deterministic exact-$K$
filling leaves $F$ unchanged.  Equation~\eqref{eq:cond-batched-bound} therefore
applies to the executed batched trajectory, rather than only to sequential
greedy selection.

To see the recurrence explicitly, write
$\Delta_t=F(S^\star)-F(S_t)$.  The singleton upper bound implies
$U_t\ge\Delta_t$, while the definition of $\alpha_t$ gives
\begin{equation}
 F(S_{t+1})-F(S_t)=\alpha_t b_tU_t/K
 \ge q_t\Delta_t.
\end{equation}
Hence $\Delta_{t+1}\le(1-q_t)\Delta_t$.  Multiplying over rounds and using
$F(S_0)\ge0$ produces Eq.~\eqref{eq:cond-batched-bound}.  The factors $q_t$
need not be assumed constant or independent; each is determined by the actual
residual matrix and selected batch at round $t$.

A complementary certificate follows from monotonicity.  Since
$F(S^\star)\le F(V)$,
\begin{equation}
 \frac{F(S)}{F(S^\star)}\ge \frac{F(S)}{F(V)}=\widehat c(S).
 \label{eq:cond-final-cert}
\end{equation}
For banks $b$ with $Z_b=F_b(V)>0$, defining
$\omega_b=Z_b/\sum_cZ_c$ and $c_b(S)=F_b(S)/Z_b$ gives the exact mixture
\begin{equation}
 \widehat c(S)=\sum_b\omega_bc_b(S),\qquad \sum_b\omega_b=1.
 \label{eq:cond-bank-mixture}
\end{equation}
This separates configured client masses from the effective facility geometry
and the saturation achieved by the final support.

The two bounds answer related but different optimization questions.
Equation~\eqref{eq:cond-batched-bound} follows the support trajectory and
compares its realized progress with an upper bound on the residual optimum.
Equation~\eqref{eq:cond-final-cert} uses only the returned support and the Full
visual set.  It is often tighter because $F(V)$ can be evaluated exactly, but
it does not describe how the solver reached $S$.  Both concern the facility
objective; the message and margin results below connect that objective to the
representation actually consumed by the decoder.

\subsection{Coverage to state distortion}

For a unit-normalized appearance client $x_m$, define
$d_m(S)=\min_{j\in S}\norm{x_m-x_j}_2$.  The Gibbs client in
Eq.~\eqref{eq:cond-aux-clients} has modal ratio
\begin{equation}
 r_m(S)=\frac{\max_{j\in S}P_m^V(j)}{P_m^V(m)}
 =\exp[-d_m(S)^2/(2\tau_v)].
 \label{eq:cond-gibbs}
\end{equation}
If $r_m(S)\ge r_0>0$, then $-\log r\le(1-r)/r_0$.  With one appearance client
per token, bank mass $\beta_V$, $p_0=\min_mP_m^V(m)$, and deficit
\begin{equation}
 D_V(S)=\frac{\beta_V}{N}\sum_mP_m^V(m)[1-r_m(S)],
\end{equation}
we obtain
\begin{equation}
 \frac1N\sum_m d_m(S)^2
 \le \frac{2\tau_v}{\beta_Vp_0r_0}D_V(S).
 \label{eq:cond-coverage-quantization}
\end{equation}
Facility deficit therefore controls average directional quantization error.

Equation~\eqref{eq:cond-gibbs} follows directly from unit normalization:
$x_m^\top x_j=1-\norm{x_m-x_j}_2^2/2$.  The row partition function cancels
between the selected maximum and the self mode.  Multiplying
$d_m^2=-2\tau_v\log r_m$ by the inequality
$-\log r\le(1-r)/r_0$ and averaging gives
\begin{align}
 \frac1N\sum_m d_m^2
 &\le\frac{2\tau_v}{Nr_0}\sum_m[1-r_m]\notag\\
 &\le\frac{2\tau_v}{\beta_Vp_0r_0}D_V(S),
\end{align}
which proves Eq.~\eqref{eq:cond-coverage-quantization}.  The constants
$p_0$ and $r_0$ make explicit when the Gibbs rows or selected modes become
diffuse; they are not absorbed into a generic saliency coefficient.

Grounded clients have an exact modal interpretation.  For client $g$, let
$e_{gi}$ denote the positive projected message energy and
$P_g^G(i)=e_{gi}/Z_g$, where $Z_g=\sum_i e_{gi}$.  If its preweight is
$\alpha_g$, the grounded facility deficit
\begin{equation}
 D_g^G=\alpha_g\left[\max_iP_g^G(i)-\max_{j\in S}P_g^G(j)\right]
\end{equation}
implies
\begin{equation}
 \max_{j\in S}e_{gj}
 =\max_i e_{gi}-\frac{Z_g}{\alpha_g}D_g^G.
 \label{eq:cond-grounded-mode}
\end{equation}
Thus grounded coverage retains high-energy carriers already accessed by a
query/head, whereas appearance coverage controls the broader boundary-state
population.

Write $h_i=\varrho_ix_i$, let $d_i^I(S)$ be the nearest normalized-state
distance among same-image representatives, and let $\lambda_i$ upper-bound the
spatial assignment bonus.  Comparing Eq.~\eqref{eq:cond-assignment} with the
nearest same-image representative gives
\begin{equation}
 \norm{x_i-x_{\pi(i)}}_2^2\le[d_i^I(S)]^2+2\lambda_i.
 \label{eq:cond-assignment-bound}
\end{equation}
For $m_i>0$, $M_j=\sum_{i\in\mathcal C_j}m_i$, and
$c_j=M_j^{-1}\sum_{i\in\mathcal C_j}m_ih_i$, completing the square yields
\begin{equation}
 \sum_{i\in\mathcal C_j}m_i\norm{h_i-u}_2^2
 =\sum_{i\in\mathcal C_j}m_i\norm{h_i-c_j}_2^2
 +M_j\norm{u-c_j}_2^2.
 \label{eq:cond-centroid-optimum}
\end{equation}
Hence $c_j$ is the unique minimum-distortion representative for a fixed
assignment.  Combining Eq.~\eqref{eq:cond-assignment-bound} with radial
decomposition bounds the pre-restoration state error by the appearance
quantization term plus explicit same-image, spatial, and radial residuals.

The global bound can be written without hiding these residuals.  Define
$\varrho_{\max}=\max_i\varrho_i$,
$m_{\max}=\max_i m_i$,
\begin{align}
 E_{\rm scale}&=\sum_i m_i(\varrho_i-\varrho_{\pi(i)})^2,\notag\\
 \Lambda&=\sum_i m_i\lambda_i,\notag\\
 \Xi_I&=\sum_i m_i\{[d_i^I(S)]^2-d_i(S)^2\}.
\end{align}
$\Xi_I$ is nonnegative and vanishes for single-image inputs.  Using
$\norm{\varrho_ix_i-\varrho_jx_j}_2^2
\le2\varrho_{\max}^2\norm{x_i-x_j}_2^2+
2(\varrho_i-\varrho_j)^2$, summing
Eq.~\eqref{eq:cond-assignment-bound}, and then applying centroid optimality
yields
\begin{equation}
 \begin{aligned}
 E_{\rm pre}
 &:=\sum_i m_i\norm{h_i-c_{\pi(i)}}_2^2\\
 &\le2\varrho_{\max}^2
 [m_{\max}N\bar d^2+\Xi_I+2\Lambda]+2E_{\rm scale},
 \end{aligned}
 \label{eq:cond-pretransport}
\end{equation}
where $\bar d^2=N^{-1}\sum_i d_i(S)^2$.  The appearance deficit bounds
$\bar d^2$ through Eq.~\eqref{eq:cond-coverage-quantization}; image
compatibility, spatial assignment, and scale mismatch remain visible.

For completeness, Eq.~\eqref{eq:cond-centroid-optimum} follows from
$h_i-u=(h_i-c_j)+(c_j-u)$.  The weighted cross term is zero because
$\sum_{i\in\mathcal C_j}m_i(h_i-c_j)=0$.  Since $M_j>0$, the remaining
$M_j\norm{u-c_j}_2^2$ is strictly positive whenever $u\ne c_j$, proving
uniqueness.

RMS restoration adds the measurable residual
\begin{align}
 E_{\rm post}&\le(1+\beta)E_{\rm pre}
 +(1+\beta^{-1})\sum_jM_j\zeta_j^2,\notag\\
 \zeta_j&=\norm{\mathcal R_j(c_j)-c_j}_2,
 \label{eq:cond-rms-bound}
\end{align}
for every $\beta>0$.

The restoration bound follows by writing
$h_i-\mathcal R_j(c_j)=(h_i-c_j)+[c_j-\mathcal R_j(c_j)]$ and applying Young's
inequality
$\norm{a+b}_2^2\le(1+\beta)\norm a_2^2+
(1+\beta^{-1})\norm b_2^2$ to every member of a cluster.  Summation produces
Eq.~\eqref{eq:cond-rms-bound}.  Centroid optimality controls the first term;
the actual scale correction $\zeta_j$ remains an explicit second term.

\subsection{Population transport and signed messages}

Before RMS restoration, Eq.~\eqref{eq:cond-transport} preserves the population
first moment exactly:
\begin{equation}
 \sum_{j\in S}M_jc_j
 =\sum_{i\in V}m_ih_i.
 \label{eq:cond-signed-moment}
\end{equation}
This identity retains signed state directions and their cancellation, despite
the nonnegative facility objective used to allocate support.

Indeed, $M_jc_j=\sum_{i\in\mathcal C_j}m_ih_i$ by construction, and the
clusters partition the source visual tokens.  Applying any linear map $W$
before summing preserves the equality:
\begin{equation}
 \sum_{j\in S}M_jWc_j=\sum_{i\in V}m_iWh_i.
 \label{eq:cond-linear-moment}
\end{equation}
The equality is vector valued.  It therefore preserves opposing directions
under the population measure, unlike an equality involving only token norms.
RMS restoration and query-dependent compact attention are subsequent
interfaces and are bounded separately below.

For a query/head $g$, let $a_{gi}$ be Full attention, $\bar a_{gj}$ compact
attention, $M_{gj}=\sum_{i\in\mathcal C_j}a_{gi}$, and
$\widetilde x_j=\mathcal N(\widetilde h_j)$.  Adding and subtracting the
transported message under Full cluster masses gives the exact decomposition
\begin{align}
 y_g-\widetilde y_g
 =&\sum_i a_{gi}W_g^V(x_i-\widetilde x_{\pi(i)})\notag\\
 &+\sum_j(M_{gj}-\bar a_{gj})W_g^V\widetilde x_j.
 \label{eq:cond-message-decomposition}
\end{align}
Consequently,
\begin{align}
 \norm{y_g-\widetilde y_g}_2
 &\le\norm{W_g^V}_2\!\left[
 \sum_i a_{gi}\norm{x_i-\widetilde x_{\pi(i)}}_2\right.\notag\\
 &\hspace{34mm}\left.+H_S\norm{M_g-\bar a_g}_1\right],
 \label{eq:cond-message-bound}
\end{align}
where $H_S=\max_j\norm{\widetilde x_j}_2$.  The bound separates transported
representation error from the compact-attention mass mismatch induced by
changed keys and sequence geometry.

The first line of Eq.~\eqref{eq:cond-message-decomposition} compares every Full
source value with the compact state of its assigned representative while
retaining Full attention weights.  The second line compares the Full attention
mass aggregated into each cluster with the mass recomputed by compact
attention.  This decomposition is exact: insert
$\sum_jM_{gj}W_g^V\widetilde x_j$ between the two messages, expand
$M_{gj}$ over $\mathcal C_j$, and group terms by assignment.  The triangle
inequality and operator norm of $W_g^V$ then yield
Eq.~\eqref{eq:cond-message-bound}.

The representation term is linked to the state distortion above.  Assume the
next RMS normalization $\mathcal N$ is locally $L_{\mathcal N}$-Lipschitz on
the two executions and recall that deployed $m_i\ge1/2$.  If
$s_g=\sum_i a_{gi}$, weighted Cauchy-Schwarz gives
\begin{align}
 R_g&:=\sum_i a_{gi}
 \norm{\mathcal N(h_i)-\mathcal N(\widetilde h_{\pi(i)})}_2\notag\\
 &\le
 \left(\sum_i\frac{a_{gi}^2}{m_i}\right)^{1/2}
 L_{\mathcal N}\sqrt{E_{\rm post}}\notag\\
 &\le\sqrt2\,L_{\mathcal N}s_g\sqrt{E_{\rm post}}.
 \label{eq:cond-state-message}
\end{align}
Together with Eqs.~\eqref{eq:cond-coverage-quantization},
\eqref{eq:cond-pretransport}, and \eqref{eq:cond-rms-bound}, this gives the
coverage-to-state-to-message chain while keeping restoration residuals
explicit.

The mass term requires no softmax linearization.  Aggregate Full visual logits
within each transport cluster,
\begin{equation}
 u^F_{gj}=\log\sum_{i\in\mathcal C_j}\exp(s^F_{gi}),
 \qquad u^S_{gj}=s^S_{gj},
\end{equation}
and retain each nonvisual key as a singleton category.  Softmax aggregation is
exact for $u_g^F$.  Since its Jacobian obeys
$\norm{J_{\rm sm}(u)v}_1\le\norm v\!_\infty$, integration along the segment
between the two category logits gives
\begin{equation}
 \norm{M_g-\bar a_g}_1\le\norm{u_g^F-u_g^S}_\infty.
\label{eq:cond-softmax-bound}
\end{equation}

To prove the Jacobian inequality, let $p=\operatorname{softmax}(u)$.  Then
$[J_{\rm sm}(u)v]_k=p_k(v_k-\mathbb E_p[v])$, and
\begin{equation}
 \norm{J_{\rm sm}(u)v}_1
 =\sum_kp_k|v_k-\mathbb E_p[v]|
 \le\norm v_\infty .
\end{equation}
The final step follows because the mean absolute deviation of values inside an
interval is at most half its range.  Integrating the Jacobian along the segment
$u(\theta)=(1-\theta)u_g^F+\theta u_g^S$ proves
Eq.~\eqref{eq:cond-softmax-bound}.  Restricting the category distribution to
visual clusters can only decrease its $\ell_1$ distance.

The grouped Full logit $u^F_{gj}$ is a log-sum-exp because several source keys
belong to one transported representative.  It is the unique scalar whose
exponential equals their total unnormalized attention mass.  Consequently the
mass term measures the combined effect of compact keys, normalization, and
position changes without approximating the Full denominator.

\paragraph{Proposition 1 (support and realization are distinct).}
Nonnegative carrier energy does not determine a signed message.  For any
$v\ne0$, the multisets $\{v,v\}$ and $\{v,-v\}$ induce the same per-token
norms and therefore the same energy-facility objective, while their signed sums
are $2v$ and $0$.  A carrier objective can determine where representational
capacity is allocated, but a second interface is necessary to determine
direction, magnitude, and cancellation.  GMC uses population transport and
native compact attention for that realization interface; their residual is
measured by Eq.~\eqref{eq:cond-message-decomposition}.

The same distinction yields a constructive directional bound.  Write a raw
message atom as $b_i=e_id_i$, where $e_i\ge0$ and $\norm{d_i}_2=1$, and let
$\pi(i)\in S$ assign it to a selected direction.  The direction-preserving
reconstruction is $\widehat b=\sum_i e_id_{\pi(i)}$.  By the triangle inequality
and weighted Cauchy-Schwarz,
\begin{align}
 \norm{\textstyle\sum_i b_i-\widehat b}_2
 &\le\sum_i e_i\norm{d_i-d_{\pi(i)}}_2\notag\\
 &\le\sqrt{2E\sum_i e_i[1-d_i^\top d_{\pi(i)}]}.
 \label{eq:cond-directional-quantization}
\end{align}
Here $E=\sum_i e_i$ is the total carrier energy.
Cosine coverage can therefore control signed-direction quantization, but it
still does not determine the population mass or the compact softmax weights.
The deployed objective uses stable nonnegative carrier energy for parallel
allocation, appearance geometry for population coverage, and the realization
interface for the signed aggregate.  The rank and direction controls in
Table~\ref{tab:cond-sensitivity} test this division without changing transport
or downstream execution.

A second exact identity separates direction from magnitude in the realized
message.  Stack weighted query/head messages into $\mathsf Y$ and
$\widetilde{\mathsf Y}$, and set
\begin{equation}
 \alpha=\frac{\norm{\widetilde{\mathsf Y}}_2}{\norm{\mathsf Y}_2},
 \qquad
 \kappa=\frac{\langle\mathsf Y,\widetilde{\mathsf Y}\rangle}
 {\norm{\mathsf Y}_2\norm{\widetilde{\mathsf Y}}_2}.
\end{equation}
Expansion of the squared difference gives
\begin{equation}
 \frac{\norm{\mathsf Y-\widetilde{\mathsf Y}}_2^2}
 {\norm{\mathsf Y}_2^2}=1+\alpha^2-2\alpha\kappa
 =(1-\alpha)^2+2\alpha(1-\kappa).
 \label{eq:cond-message-polar}
\end{equation}
Thus relative signed-message error is at least $|1-\alpha|$ and at least
$\sqrt{2\alpha(1-\kappa)}$.  Matching amplitude alone cannot ensure direction,
and matching direction alone cannot ensure amplitude.  The grouped-mass term
in Eq.~\eqref{eq:cond-message-bound} is independent of both because it concerns
how compact attention weights the representatives.

\subsection{Visual innovation and candidate margins}

At answer position $t$ under a shared history $h$, let $z_t^F(h)$,
$z_t^S(h)$, and $z_t^0(h)$ be the Full, compact, and visual-null logits.  For
$C=\operatorname{Top}_{32}(z_t^F)\cup\operatorname{Top}_{32}(z_t^0)$ and
$H_C=I-|C|^{-1}\mathbf1\mathbf1^\top$, visual innovation is the centered
anti-prior displacement
\begin{equation}
 \nu_t^E(h)=H_C[z_{t,C}^E(h)-z_{t,C}^0(h)],\qquad E\in\{F,S\}.
 \label{eq:cond-innovation}
\end{equation}
The reported VIE is a fixed Full/null-weighted relative norm of
$\nu_t^F-\nu_t^S$; it is a diagnostic and does not enter support selection.

Let $e_\ell=\norm{T_\ell^F-T_\ell^S}_2$ be text-state error and
$\epsilon_\ell=\norm{M_\ell^F-M_\ell^S}_2$ message error.  If block $\ell$
is locally Lipschitz in these arguments,
\begin{equation}
 e_{\ell+1}\le a_\ell e_\ell+b_\ell\epsilon_\ell.
\end{equation}
Writing $A_{u:v}=\prod_{r=u}^{v}a_r$ and using an output constant $L_o$ gives
\begin{align}
 d_{t,C}&:=\norm{H_C(z_{t,C}^F-z_{t,C}^S)}_2,\notag\\
 d_{t,C}&\le L_o\!\left[e_pA_{p:L-1}+
 \sum_{\ell=p}^{L-1}b_\ell\epsilon_\ell A_{\ell+1:L-1}\right].
 \label{eq:cond-innovation-bound}
\end{align}
Thus the signed boundary-message bound propagates to centered candidate logits
through the remaining native decoder blocks.

Equation~\eqref{eq:cond-innovation-bound} is obtained by repeatedly substituting
the one-block recurrence.  In particular,
\begin{equation}
 e_L\le e_p\prod_{r=p}^{L-1}a_r+
 \sum_{\ell=p}^{L-1}b_\ell\epsilon_\ell
 \prod_{r=\ell+1}^{L-1}a_r,
 \label{eq:cond-unrolled-state}
\end{equation}
where an empty product equals one.  Centering is a nonexpansive orthogonal
projection, so composing the final normalization and language head with its
local constant $L_o$ gives Eq.~\eqref{eq:cond-innovation-bound}.  The
coefficients expose where an upstream perturbation enters and how many native
decoder blocks can subsequently amplify it.

The shared-history condition specifies the scope of this local comparison.
For the no-cache recurrence in Eq.~\eqref{eq:cond-no-cache-rollout}, both
executions run the same dense prefix blocks when conditioned on the same text
history, hence $e_p=0$ before the compact intervention.  In cached execution,
$e_p=0$ at the prompt-terminal decision; later shared-history probes may see a
nonzero boundary term because one cache contains gathered compact prompt K/V.
For the models' own rollouts, let $h_t^F=(q,y_{<t}^F)$ and
$h_t^S=(q,y_{<t}^S)$.  The boundary discrepancy decomposes as
\begin{align}
 \norm{T_p^F(h_t^F)-T_p^S(h_t^S)}_2
 \le{}&\norm{T_p^F(h_t^F)-T_p^S(h_t^F)}_2\notag\\
 &+\norm{T_p^S(h_t^F)-T_p^S(h_t^S)}_2.
 \label{eq:cond-rollout-decomposition}
\end{align}
The first term is compression under a shared history; the second is the effect
of a previously diverged generated prefix.  Prompt-terminal VIE measures the
first term before history divergence, while full task and hallucination metrics
evaluate each compressed model's complete own-history rollout.

For candidates $a,b\in C$, $(e_a-e_b)^\top\mathbf1=0$, so Cauchy-Schwarz
implies
\begin{equation}
 |(z_a^S-z_b^S)-(z_a^F-z_b^F)|\le\sqrt2\,d_{t,C}.
 \label{eq:cond-margin-bound}
\end{equation}
If Full's margin $\gamma_{ab}=z_a^F-z_b^F$ is positive and
$d_{t,C}<\gamma_{ab}/\sqrt2$, compact execution preserves the ordering of
that pair.  More generally, for Full winner $a_C^F$, minimum represented margin
$\Gamma_C^F$, and any $\eta>0$,
\begin{align}
 \Pr(a^S\ne a^F)\le{}&
 \Pr(a^S\notin C)+\Pr(d_{t,C}>\eta)\notag\\
 &+\Pr(\Gamma_C^F\le\sqrt2\eta).
 \label{eq:cond-flip-risk}
\end{align}
To prove Eq.~\eqref{eq:cond-margin-bound}, observe that pairwise logit
differences are invariant to centering:
$(e_a-e_b)^\top H_C=(e_a-e_b)^\top$.  Since
$\norm{e_a-e_b}_2=\sqrt2$, Cauchy-Schwarz bounds the displacement of every
represented pair by $\sqrt2d_{t,C}$.  If the Full winner's minimum margin in
$C$ exceeds this displacement, it remains the winner within $C$.  A
full-vocabulary winner change can then occur only if the compact winner escapes
$C$, distortion exceeds a chosen threshold $\eta$, or Full's represented
margin lies below $\sqrt2\eta$.  Taking the union bound gives
Eq.~\eqref{eq:cond-flip-risk}.

The prospective set $C_t^{\rm AP}$ in
Eq.~\eqref{eq:cond-vie-candidates} is frozen from Full and visual-null logits
before compact logits are examined.  It tests whether image-induced and
language-prior candidates jointly contain the eventual compact winner.  The
finite no-flip check uses the separate set
$C_x^{\rm win}=\operatorname{Top}_{32}(z_x^F)\cup
\operatorname{Top}_{32}(z_x^S)$, which contains the Full top two and both
observed winners.  Its $\Gamma_x^F$ is therefore Full's vocabulary-wide
top-two margin, and Eq.~\eqref{eq:cond-margin-bound} gives
\begin{equation}
 \rho_x=\frac{\sqrt2
 \norm{H_{C_x^{\rm win}}(z_x^F-z_x^S)}_2}{\Gamma_x^F}<1
 \quad\Longrightarrow\quad a_x^S=a_x^F.
 \label{eq:cond-observed-certificate}
\end{equation}
The implication is deterministic for each condition.  Bootstrap intervals
describe how frequently it applies in the evaluated population.  The observed
21.6\% coverage with zero violations is therefore an empirical margin
certificate derived from the exact output logits, while
Eq.~\eqref{eq:cond-innovation-bound} remains the analytical interface bound
from upstream message perturbations.

Combining Eqs.~\eqref{eq:cond-message-bound},
\eqref{eq:cond-innovation-bound}, and \eqref{eq:cond-margin-bound} gives an
explicit sufficient budget for a represented pair:
\begin{equation}
 \sqrt2L_o\!\left[e_pA_{p:L-1}+
 \sum_{\ell=p}^{L-1}b_\ell\epsilon_\ell A_{\ell+1:L-1}\right]
 <\gamma_{ab}.
 \label{eq:cond-interface-budget}
\end{equation}
Support coverage reduces the assignment component of $\epsilon_p$; transport
minimizes the fixed-assignment state term; original positions and compact keys
control grouped attention-mass mismatch; the remaining blocks propagate the
realized perturbation.  The bound therefore assigns distinct roles to support
allocation and compact representation while ending at the candidate ranking
used by autoregressive decoding.

\paragraph{Composing the two interfaces.}
The preceding results can be combined into one sufficient condition without
identifying nonnegative facility energy with a signed attention message.  For
query/head $g$ at layer $\ell$, define
\begin{align}
 R_{\ell g}&=\sqrt2L_{\mathcal N}s_{\ell g}
 \sqrt{E_{\rm post}^{\ell}},\notag\\
 \delta_{\ell g}&=\norm{M_{\ell g}-\bar a_{\ell g}}_1,\notag\\
 B_{\ell g}&=\norm{W_{\ell g}^V}_2
 [R_{\ell g}+H_{\ell S}\delta_{\ell g}] .
 \label{eq:cond-local-message-budget}
\end{align}
Equation~\eqref{eq:cond-state-message} and the grouped-mass term in
Eq.~\eqref{eq:cond-message-bound} imply
$\norm{y_{\ell g}-\widetilde y_{\ell g}}_2\le B_{\ell g}$.
If the layer stacks messages using nonnegative evaluation weights
$\eta_{\ell g}$, then
\begin{equation}
 \epsilon_\ell\le
 \left(\sum_g\eta_{\ell g}B_{\ell g}^2\right)^{1/2}
 =:\mathcal B_\ell .
 \label{eq:cond-stacked-message-budget}
\end{equation}
Here $E_{\rm post}^{\ell}$ is bounded by appearance deficit, same-image
restriction, spatial assignment, radial mismatch, and restoration through
Eqs.~\eqref{eq:cond-coverage-quantization}-\eqref{eq:cond-rms-bound};
$\delta_{\ell g}$ is the separately measured compact-attention mass mismatch.

Substitution into Eq.~\eqref{eq:cond-interface-budget} yields
\begin{equation}
 \sqrt2L_o\!\left[e_pA_{p:L-1}+
 \sum_{\ell=p}^{L-1}b_\ell\mathcal B_\ell
 A_{\ell+1:L-1}\right]<\gamma_{ab}.
 \label{eq:cond-composed-budget}
\end{equation}
Whenever Eq.~\eqref{eq:cond-composed-budget} holds, the represented ordering
between $a$ and $b$ cannot change.  The inequality exhibits the two method
interfaces explicitly.  Facility coverage and assignment limit the first part
of $\mathcal B_\ell$; transport is the fixed-assignment optimum before RMS
restoration; compact keys and source positions determine the second part; the
native suffix contributes the propagation factors.

The result is local because transformer blocks and RMS normalization are
nonlinear, and its constants depend on the evaluated states.  Its useful
content is the decomposition: each residual has a measurable counterpart and
no signed or positional error is absorbed into facility coverage.  The
empirical output certificate in Eq.~\eqref{eq:cond-observed-certificate}
evaluates the final left-hand effect directly through logits and the realized
Full margin.  The two forms therefore provide complementary resolutions of the
same event: Eq.~\eqref{eq:cond-composed-budget} traces how compression error can
propagate, while $\rho_x$ tests the resulting winner stability without
estimating intermediate Lipschitz constants.

\section{Reproducibility Summary}

\paragraph{Models and software.}
Experiments use Python
3.10.19, PyTorch 2.8.0 with CUDA 12.8, Transformers 4.51.3, and LMMS-Eval
0.3.5.  Controlled Full, GMC, and comparator rows share the checkpoint-saved
image processor, prompt construction, batch size one, greedy generation, and
task-specific output limit.  Qwen and LLaVA decoder execution uses
\textsc{bfloat16} and PyTorch SDPA.  Message construction, residual gains,
population weights, and scatter accumulations use FP32 before transported
states are cast to the residual-stream dtype.

\paragraph{Inputs and generation.}
The fixed-token Qwen protocol bicubically resizes each image to
$1008\times1008$, producing $N=1{,}296$ merged visual tokens.  LLaVA uses its
frozen processor and $N=576$.  The document deployment resizes the paired Full
and GMC inputs to $3528\times3528$ with the same bicubic rule and processor
ceiling, producing $N=15{,}876$.  Table~\ref{tab:cond-eval} records dataset
splits, sample counts, answer formats, and generation limits.  TextVQA and
DocVQA request a single word or phrase; ChartQA requests a single word; POPE,
MME, and AMBER use the common short-answer suffix; HallusionBench requests
yes/no only; CHAIR uses the detailed-caption prompt.  Every compressed row
uses the same model EOS behavior as Full.

\paragraph{Determinism and configuration.}
Python, NumPy, and PyTorch seeds are 0, 1234, and 1234.  Message projection uses
Gaussian-QR seed zero; long-prefix appearance projection uses Rademacher seed
104729.  Farthest-first initialization and all equal-score choices use stable
source-token order.  A requested multi-image budget smaller than the number of
images expands to one representative per image; a budget above the source
length clamps to $N$; zero-gain completion preserves the facility value while
returning exactly $K$ positions.  Table~\ref{tab:cond-config} contains the
shared bank masses, temperatures, history depth, responsibility batch,
assignment gate, transport strength, and bounded long-prefix settings.  These
settings are common across fixed-token tasks and transfer to LLaVA without
architecture-specific retuning.

\paragraph{Boundary and sequence semantics.}
H1, H2, and L16 compact before blocks 1, 2, and 16.  Their completed historical
views are $\{0\}$, $\{0,1\}$, and $\{14,15\}$; the current next-block view is
stored separately.  Client construction uses the normalization, M-RoPE
coordinates, masks, and Q/K/V projections of the actual split boundary.
Selected source tokens represent themselves during assignment, discarded
tokens cannot cross image identities, and every compact representative retains
its complete source multimodal position ID.  Text states and text positions are
unchanged by compaction.  The reported final $K$ is checked after physical
deletion and is the number of visual states entering every post-boundary block;
token-layer work is then computed from Eq.~\eqref{eq:cond-work}.

\paragraph{Evaluation and uncertainty.}
Full and compressed predictions are joined by exact sample identifier before
metric computation.  TextVQA, ChartQA, POPE, AMBER, MMBench, CHAIR, and DocVQA
use their task-native evaluator; MME reconstructs category-pair totals; linked
HallusionBench records move together.  Paired intervals resample these native
units and use the same sampled indices for every method.  Factorial contrasts
are recomputed inside each replicate, and equal-task macros normalize MME
before averaging.  The general-quality and low-budget faithfulness families apply Holm correction independently, as described in the statistical protocol.  Qualitative examples are selected from stored model predictions
and use the corresponding source images; no OCR system, detector, or verifier
contributes to an answer.

\paragraph{Timing and memory.}
The H800 system table uses one physical allocation, the common SDPA backend,
batch size one, one warm-up pass, synchronized measurement boundaries, and
identical examples.  Its elapsed time starts from processed model inputs and
ends after generation, including message extraction, selection, transport,
physical compaction, and decoder execution.  The cached document path uses one
A800 with fixed $N=15{,}876$, matched sample offsets, and a 16-token generation
horizon.  Prompt-KV bytes are derived from the realized compact sequence and
number of cached layers.  Peak allocation is reset before each measured run and
includes weights, dense-prefix activations, client buffers, cache, and allocator
reserve; it is therefore reported separately from prompt-KV reduction.

\paragraph{Inference dependencies.}
GMC accesses internal boundary states, native Q/K/V projections, the
question-to-visual attention slice, and multimodal position IDs.  It performs
no parameter update or gradient computation and invokes no task label, learned
scorer, detector, OCR engine, auxiliary verifier, or second VLM at inference.
Each result record stores the requested and realized budget, source and compact
sequence lengths, pruning boundary, selected source indices, original retained
positions, bank objectives, work count, generated text, and task score.  These
fields are sufficient to recompute final support saturation, verify the
physical token budget, align paired outputs, and reproduce the aggregate tables
without another model forward.

\paragraph{Controlled comparator configuration.}
The MMTok execution uses released revision, with text-visual temperature $0.01$,
visual-visual temperature $0.2$, and mixture coefficient $0.5$.  Its adapter
receives the same frozen Qwen checkpoint, image tensors, prompt strings,
generation limits, and physical target $K$ as GMC.  VisionZip and FastV-style
use their released selection equations inside the common Qwen split-forward
runner.  The H800 timing rows force SDPA for all methods and count any method
specific selection and compaction before generation.  Same-path controls use
the common GMC split boundary and change only the returned support indices;
the support is then passed through the identical image-compatibility,
assignment, transport, position, and decoder code.

\paragraph{Reproduction sequence.}
A benchmark run first writes the shared Full prediction keyed by the native
sample identifier.  A compressed run processes the identical record order,
constructs the coreset once from the image-prompt prefix, verifies the realized
cardinality, and generates its answer under
Eq.~\eqref{eq:cond-no-cache-rollout} or
Eq.~\eqref{eq:cond-cached-rollout}.  Task adapters then normalize model output
according to the corresponding evaluator and join Full, GMC, and comparator
records by identifier.  Native task scores are computed before Full-relative
retention or equal-task averaging.  Finally, paired bootstrap indices are
sampled from the joined population, so missing or duplicated records cannot
silently alter a contrast.

\paragraph{Numerical checks.}
Before post-boundary execution, the runner checks that selected indices are
unique, in range, image compatible, and equal in number to the realized budget.
The stable compaction map must retain every text position and exactly the
selected visual positions.  Each transported cluster has positive mass because
selected tokens self-assign and all population weights are positive.  FP32
centroid accumulation is converted only after RMS restoration; finite-value
checks cover client normalization, responsibility gains, population masses,
transported states, and compact logits.  The $K=N$ endpoint bypasses selection
and returns the unmodified sequence, providing an identity check for every
supported backbone and boundary.

\paragraph{Aggregate reconstruction.}
The fixed-$K$ table reads native scores from the shared prediction sets and
computes each taskwise retention against its single Full row.  Equal-work
budgets are integer solutions closest to the target
$W(p,K)$ and are displayed with their realized work, so rounding remains
visible.  Runtime summaries join quality and timing by method, boundary,
source length, budget, generation horizon, backend, and hardware.  Component
timings partition message construction, selection, transport, and compaction;
their sum is reported independently of whole-run latency.  This organization
keeps absolute quality, compact capacity, decoder work, persistent KV memory,
temporary peak allocation, and elapsed time as separate measured quantities.

\paragraph{Task adapters.}
TextVQA uses the benchmark answer list and its case/punctuation normalization;
ChartQA applies the released relaxed numerical matcher; DocVQA computes ANLS
from the complete reference set.  POPE reports binary accuracy and F1 after its
yes/no normalizer, while MME aggregates paired perception questions into the
official category total.  HallusionBench uses official set clusters and its
accuracy aggregation.  AMBER and CHAIR retain the same output horizon and
normalization for Full and every compressed method.  Multiple-choice MMBench
predictions are reduced to the selected option letter before scoring.  The
adapter is selected from the task name rather than model output, so a compressed
answer cannot alter its scoring rule.

\paragraph{Document deployment sampling.}
The cached deployment quality set is a deterministic three-way contiguous shard
of 300 DocVQA validation questions.  Full, H1, and H2 share its membership,
image preprocessing, prompt, and generation horizon.  Timing repetitions
preserve these offsets and separate warm-up from measured passes.  The complete
5,349-question DocVQA validation set is used for the message-path rate
experiment in Table~\ref{tab:cond-message-path}; the 300-question set is used
for the paired quality-systems operating points in
Tables~\ref{tab:cond-runtime} and \ref{tab:cond-overhead-cached}.

\paragraph{Support and output records.}
Support indices are stored in source-grid order together with image identity
and retained multimodal coordinates.  Per-bank numerators and denominators
record grounded, appearance, and spatial saturation before transport; native
signed-message quantities are computed after compact attention.  Prediction
records store both raw generated text and evaluator-normalized text.  Support,
representation, and answer metrics can therefore be recomputed at their
corresponding interfaces while preserving one common sample key.

\end{document}